\documentclass{article} 
\usepackage{style/arxiv}

\usepackage{url}
\usepackage{import}
\usepackage{multirow}
\usepackage{subcaption}
\usepackage{graphicx}
\usepackage{amsmath}
\usepackage[table]{xcolor}
\usepackage{soul}
\usepackage{tabularx}
\usepackage{booktabs}

\usepackage{lipsum}

\usepackage[pagebackref=true,breaklinks=true,letterpaper=true,colorlinks,bookmarks=false]{hyperref}

\begin{document}

\title{Gaze Authentication: Factors Influencing Authentication Performance}

\author{
    Dillon~Lohr\textsuperscript{1}, 
    Michael J. Proulx\textsuperscript{1}, 
    Mehedi Hasan Raju\textsuperscript{2},  
    Oleg V. Komogortsev\textsuperscript{1,2}\\
    \textsuperscript{1}Meta Reality Labs Research, Redmond, WA, USA, 
    \textsuperscript{2}Texas State University, San Marcos, Texas, USA \\
    {\tt\small dlohr@meta.com, michaelproulx@meta.com, m.raju@txstate.edu, ok@txstate.edu}
}
\date{}

\maketitle
\begin{abstract}

This paper examines the key factors that influence the performance of state-of-the-art gaze-based authentication. Experiments were conducted on a large-scale, in-house dataset comprising 8,849 subjects collected with Meta Quest Pro equivalent hardware running a video oculography-driven gaze estimation pipeline at 72~Hz. State of the neural network architecture was employed to study the influence of the following factors on authentication performance: eye tracking signal quality, various aspects of eye tracking calibration, and simple filtering on estimated raw gaze. This report provides performance results and their analysis.

\end{abstract}


\section{INTRODUCTION}

The growing use of extended reality (XR) devices brings both opportunities and security challenges.
It is important to ensure continuous and secure user authentication on these devices \cite{lohr2024baseline}.
Many XR headsets now include built-in eye-tracking (ET) technologies for foveated rendering \cite{adhanom2023eye, foveated_rendering,foveated_rendering2}
and gaze-based interaction \cite{piumsomboon2017exploring, raju2025interaction}. 
This native ET integration facilitates the development of an authentication system for continuous user authentication \cite{lohr2025ocular}.

Gaze-based authentication (GA) systems use person-specific information embedded in ET signals to verify user identity based on eye movements \cite{komogortsev2013biometric, rigas_emb_review, raju2024signal_noise}.
These systems operate on the principle that authentication can be achieved by exploiting the distinctiveness and repeatability of an individual's physiological, behavioral traits, or, in this case, a psychophysiological trait (eye movement) \cite{Jain2025, jain2007handbook}. 
Eye movement has demonstrated promise as a user authentication modality over the past couple of decades \cite{Kasprowski2004,rigas_emb_review,deepeyedentification,lohr_ekyt}.
Its resilience to spoofing \cite{rigas2015,raju2022iris} and strong user-specific signal characteristics make it particularly attractive for authentication systems \cite{eberz2015preventing, komogortsev2015}.

Prior study reported that a best-case result of 2.4\% false rejection rate (FRR) at a 1-in-50,000 false acceptance rate (FAR) when using 20 seconds of gaze data during a random saccade (jumping dot) task in \cite{lohr2024baseline}. 
To the best of our knowledge, it was the first report to demonstrate high user authentication accuracy using gaze data alone.
The achieved authentication performance exceeds the FIDO benchmark \cite{FIDO2020} requirement even with consumer-facing XR hardware.
This milestone underscores the need for further in-depth research in this domain.
The main objective of this study is not to outperform existing benchmarks but to investigate the key factors within the GA pipeline that influence authentication accuracy.
Nevertheless, any improvement in performance arising from these insights would be an added advantage.

\textit{What are these key factors?} 
First, calibration of gaze in an ET pipeline \cite{nystrom2025fundamentals} is an important factor because it establishes the mapping between raw eye tracker signals and gaze positions in the user’s visual scene.
This process adapts the system to individual differences in anatomy and device placement, both essential for robust and generalizable authentication performance.
A specific aspect we examine is how calibration data from different calibration target depths affects calibration performance.

Second, prior study \cite{lohr2024baseline} emphasized the importance of ET signal quality.
In this study, we revisit this factor in the context of an improved gaze-estimation pipeline, examining whether signal quality improvements translate to better authentication outcomes.
Third, while most of the studies in this domain employed estimates of the visual axis, some approaches explore calibration-free optical axis instead \cite{lohr2025ocular}.
Based on prior findings, we investigate whether prioritizing one axis over the other or combining them yields measurable benefits for authentication tasks.
Additionally, prior research \cite{raju2024signal_noise} demonstrated that authentication performance improved with filtering. 
We extend this line of inquiry to evaluate how basic filtering steps affect performance within an updated pipeline presented in this work.
Moreover, calibration parameters, training depth are additional factors investigated in this work.

This study conducts a comprehensive parameter evaluation to systematically examine how these key factors affect the authentication performance of our state-of-the-art approach \cite{lohr2024baseline}. 
Rather than conducting controlled ablation studies that isolate individual factors, our evaluation approach assesses performance across diverse configurations to understand their practical impact on system performance under real-world conditions. 
This methodology differs from traditional ablation studies, which typically hold all factors constant while varying one element at a time. 
Instead, our work explores how multiple configurations interact to inform deployment best practices for gaze-based biometric systems.

The following research questions will guide our investigation:

\textit{\textbf{(RQ1)} What effect do various aspects of gaze calibration have on the resulting authentication accuracy?}

\textit{\textbf{(RQ2)} What effect do different calibration parameters have during training on authentication performance?}  

\textit{\textbf{(RQ3)} How does ET signal quality, specifically spatial accuracy and precision, affect authentication accuracy?} 

\textit{\textbf{(RQ4)} How performance provided by the visual axis compare to the performance of the optical axis?} 

\textit{\textbf{(RQ5)} What's the impact of training depth?} 

\textit{\textbf{(RQ6)} How simple filtering affects authentication performance?} 

The investigation of those questions is conducted on a very large dataset of 8,849 people, something that was not done before.

\section{PRIOR WORK}

Kasprowski and Ober introduced eye movements as a biometric modality for user authentication in \cite{Kasprowski2004}.
Eye movements from nine subjects were recorded with a jumping point-of-light stimulus. Biometric features were extracted using spectrum analysis and classified using different classifiers.
One of the primary reasons for the sustained interest in eye movement data is its inherent resistance to imitation and spoofing \cite{rigas2015,raju2022iris}. 
Since eye movements are driven by both cognitive processes and physiological traits, they are exceptionally difficult to replicate with precision \cite{deepeyedentificationlive, lohr_ekyt, lohr2020_metric, lohr2020_vr_ex_abs, deepeyedentification, rigas2015, raju2022iris, komogortsev2015}.

Based on their pioneering work, numerous studies have contributed to the advancement of GA systems, exploring various methodologies and applications.
For instance, Bednarik et al. \cite{bednarik2005eye} analyzed static and dynamic eye features from 12 subjects using a static cross stimulus. After dimensionality reduction (FFT, PCA, or both), k-NN classification yielded 40-50\% accuracy for dynamic features and up to 90\% for static features.
Komogortsev et al. further advanced the field by proposing the Oculomotor Plant Mathematical Model (OPMM) and used it for authentication based on Oculomotor Plant Characteristics (OPC) \cite{komogortsev2010biometric, komogortsev2012biometric}.
Holland and Komogortsev \cite{holland2011} also analyzed primitive eye movement features during text reading, including fixation duration and position, and saccadic amplitude, duration, and velocity.
Later, they expanded their analysis to Complex Oculomotor Behavior (COB) features, including saccadic dysmetria, compound saccades, dynamic overshoot, and express saccades using a jumping point-of-light stimulus with 32 subjects \cite{komogortsev2013biometric}.

With the advancement of the research domain, machine learning and deep learning have become central to gaze-based authentication systems. 
Two main approaches have emerged. The first uses pre-extracted, handcrafted features from eye movement data, such as fixation durations, saccade amplitudes \cite{lohr2020_metric, george2016score}.
The second adopts end-to-end models that learn discriminative representations directly from raw eye-tracking data, using architectures such as convolutional and recurrent neural networks to capture temporal and spatial gaze patterns without manual feature engineering \cite{Jia2018, lohr_eky, deepeyedentification, deepeyedentificationlive, lohr_ekyt}.


Prior research \cite{raju2024temporal} has demonstrated that a sampling rate as low as 100~Hz provides high reliability and strong biometric performance. 
This finding is particularly relevant for XR environments, where devices like the Meta Quest Pro operate at similar sampling rates (e.g., 90~Hz), enabling the investigation of GA under more naturalistic and unconstrained user interactions.
This line of research has shown that even consumer-facing eye trackers embedded in head-mounted displays can achieve competitive authentication performance \cite{lohr2018implementation, lohr2020eye, lohr2023demonstrating, raju2025spw}.
The latest addition to this domain is our prior work, where we conducted a breadth-first investigation on gaze-driven authentication performance in XR with a very large dataset (nearly 10000 people).
It is important to mention that an important milestone was achieved in that study, with authentication performance reaching remarkably low equal error rates. 
This performance was 0.04\% in best cases with an FRR of 2.4\% at a 1-in-50,000 FAR when using 20 seconds of gaze only \cite{lohr2024baseline}.

The most recent work in this area has employed state-of-the-art network architecture trained on datasets with subject pools of a scale not currently available. 
Consequently, our goal in this study is not to improve upon existing benchmarks but rather to investigate the factors within the current state-of-the-art authentication pipeline that may impact GA accuracy.

Finally, numerous studies continue to enhance this field, exploring fusion techniques, task-specific feature robustness, and deep learning advancements \cite{holland2013, komogortsev2014biometrics, nigam2015ocular, zhang2016biometrics, kasprowski2018fusion, li2018biometric, brasil2020eye, katsini2020role, liao2022exploring, raju2024signal_noise, grandhi2025evaluating, EmMixformer2025}. 
Several surveys have synthesized these developments, highlighting GA's potential alongside its challenges and emerging research directions \cite{galdi2016eye, rigas_emb_review, zhang2015survey}.

\section{METHODOLOGY}

\subsection{Dataset}

We have used the internal GazePro dataset of gaze signals \cite{lohr2024baseline}.
We had access to gaze data from 8,849 people (out of a total of 9,202) on the version-3 hardware revision using the latest gaze estimation model.
The data was collected at 72~Hz with ET signal quality similar to Meta Quest Pro \cite{wei2023preliminary,mqp}
The appearance-based gaze estimation method used for the device that captured the data was similar to Meta's Project Aria \cite{aria} Eye Tracking approach\footnote{\href{https://github.com/facebookresearch/projectaria_eyetracking}{https://github.com/facebookresearch/projectaria\_eyetracking}}.

GazePro includes 9,202 participants: 3,673 identified as male, 5,376 as female, and 153 as neither. 
Participants' ages ranged from 13 to 88 years, with a median age of 33. 
All individuals had normal or corrected-to-normal vision, with 3,223 wearing glasses or contact lenses and 5979 wearing no corrective lenses.
A total of 2,032 participants were reported to be wearing some form of eye makeup, including eyelash extensions, heavy eyeshadow, and/or mascara. 
In terms of eye color, 5,614 participants had brown eyes, followed by 1,759 with blue, 1,113 with hazel, 701 with green, 13 with amber, and 2 with violet eyes.
\color{black}

Participants were not required to use a chin rest while data were being collected.
Each person performed up to 8 eye-tracking tasks, two of which were used to fit calibration parameters for estimating the visual axis.  
By ``visual axis'', we mean gaze directions that were estimated by Quest Pro gaze model and additionally had a user-specific calibration applied.
By ``optical axis'', we mean gaze directions that have not undergone calibration — that is, the original estimation of gaze comes from an appearance-driven gaze estimation pipeline.

The experimental protocol consisted of a series of tasks, including random saccades, smooth pursuit, and vestibulo-ocular reflex exercises. 
Each task featured a single focal target to encourage sustained visual attention. 
The entire session lasted approximately 20 minutes. Several tasks were adapted from the procedures described by Lohr et al. \cite{gazebasevr}.
Further details about the data collection procedure are available in \cite{lohr2025ocular,lohr2024baseline}

\subsection{Data Processing}
\label{sec:data-processing}

Prior to network training, all recordings in the dataset underwent a series of preprocessing pipelines. 
First of all, horizontal and vertical velocity channels were derived from the raw signals using a Savitzky-Golay filter \cite{savitzkyGolayM} with a window size of 7 and a polynomial order of 2 \cite{friedman2017method}. 
The resulting velocity signals were segmented into non-overlapping 5-second windows (corresponding to 360 samples --- 5 seconds @ 72~Hz) employing a rolling window strategy. 
For subsequent analyses, multiple contiguous 5-second segments were concatenated to yield longer sequences of 10 and 20 seconds in duration. 
To attenuate high-frequency noise, velocity magnitudes were constrained within a range of ±1000°/s. 
Thereafter, the clamped velocities were standardized to zero mean and unit variance according to the transformation $(x - \mu)/\sigma$, where $\mu$ and $\sigma$ denote the mean and standard deviation estimated exclusively from the training data. 
Missing values (NaNs) were imputed with zero to ensure numerical stability. 
Further methodological details regarding the preprocessing procedure are available in \cite{lohr_ekyt}.

\subsection{Network Architecture}

We have employed Eye Know You Too (EKYT) \cite{lohr_ekyt}, a state-of-the-art user authentication model based on DenseNet architecture \cite{densenet}.
EKYT has demonstrated outstanding performance in biometric authentication, especially with large volumes of data collected at 72~Hz, achieving an equal error rate (EER) of 0.04\% by utilizing both visual and optical axis estimations \cite{lohr2024baseline}.
The architecture consists of eight convolutional layers, with the feature maps from each layer concatenated with those from all previous layers before being passed to the next convolutional layer.
The resulting concatenated feature maps are flattened, undergo global average pooling, and fed into a fully connected layer to generate 128-dimensional embeddings.
For a detailed description of the network architecture, see \cite{lohr_ekyt}.

\begin{figure*}[ht!]
    \centering        
    \includegraphics[width=0.65\columnwidth]{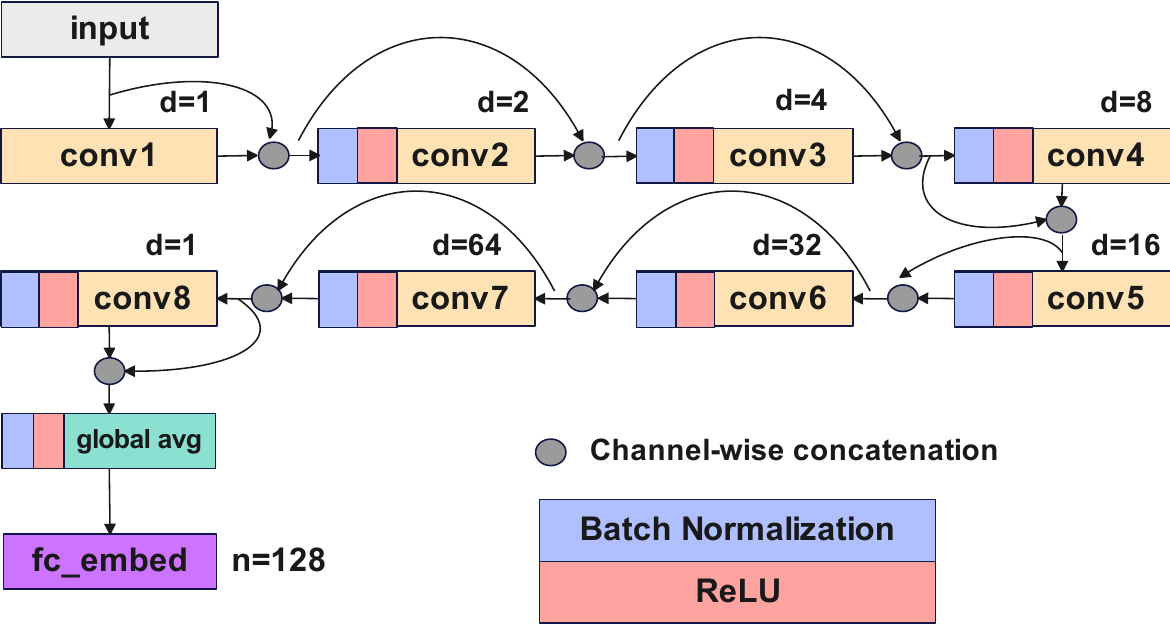}
    \caption{Overview of the embedding model’s DenseNet-based network architecture \cite{lohr_ekyt}.  The input is 360 time steps (5 seconds @ 72 Hz), where each time step has 4 or 8 features (yaw/pitch velocity $\times$ left/right eye $\times$ optical/visual axis), depending on whether we use one or both axes. The output is a 128-dimensional embedding.}
    \label{fig:network}
\end{figure*}

\subsection{Model Training}

We had 6,134 training subjects and 1,983 testing subjects.  
Of these, 143 training subjects were excluded because they had fewer than 64 samples (320 seconds of data) across all tasks and calibrations.

In the training process, we closely adhered to the training methodology used for \cite{lohr2024baseline}. 
The model is trained using the multi-similarity (MS) loss function~\cite{Wang2019}, defined as

\begin{align}
\mathcal{L}_{\text{MS}} = \frac{1}{m} \sum_{i=1}^{m} \Bigg[ 
    &\frac{1}{\alpha} \log \left( 1 + \sum_{k \in \mathcal{P}_i} \exp \left[ -\alpha (S_{ik} - \lambda) \right] \right) \nonumber \\
  +\; &\frac{1}{\beta} \log \left( 1 + \sum_{k \in \mathcal{N}_i} \exp \left[ \beta (S_{ik} - \lambda) \right] \right) 
\Bigg],
\end{align}

where $m = 256$ is the minibatch size; $\alpha = 2.0$, $\beta = 50.0$, and $\lambda = 0.5$ are hyperparameters; $\mathcal{P}_i$ and $\mathcal{N}_i$ denote the sets of mined positive and negative pairs for the anchor sample $x_i$; and $S_{ik}$ represents the cosine similarity between samples $x_i$ and $x_k$. This formulation of MS loss implicitly incorporates an online pair mining strategy with an additional hyperparameter $\epsilon = 0.1$. Further details on MS loss can be found in~\cite{Wang2019}.

Each minibatch is constructed by randomly selecting 16 unique users from the training set and sampling 16 distinct examples per user, yielding $16 \times 16 = 256$ samples per minibatch. 
One training epoch comprises as many such minibatches as necessary to approximate the total number of unique samples in the training set; however, individual samples are not guaranteed to be included in every epoch. 

Model training is conducted over 100 epochs using the Adam optimizer~\cite{adam}, combined with a one-cycle cosine annealing learning rate schedule~\cite{smith2019super}. 
The learning rate is initialized at $10^{-4}$, gradually increases to a peak of $10^{-2}$ during the initial 30 epochs, and subsequently decays to a minimum of $10^{-7}$ over the remaining 70 epochs.

\subsection{Model Evaluation}

During the evaluation phase, we select two mutually exclusive subsets of ET recordings: one for enrollment and another for verification. 
To prevent participant-specific bias, each participant contributes at most one recording to each subset, and no recording is reused across both. 
The recordings used for evaluation are processed using the same preprocessing pipeline described in Section~\ref{sec:data-processing}, ensuring consistency with the training data.

The gaze estimation pipeline employed in GazePro continuously provides gaze coordinates for all sampled time points, including intervals corresponding to eye blinks and periods of complete eye closure. 
As a result, we do not impose the minimum gaze validity threshold that was required in prior studies~\cite{lohr_ekyt}, since it is redundant in this setting.
For each eye-tracking recording, we extract the first $n$ valid 5-second segments and process each segment independently through the trained model to obtain an embedding representation.
These $n$ embeddings are then averaged to produce a single centroid embedding per recording. 

This procedure is applied identically to both the enrollment and verification recordings, resulting in two disjoint sets of centroid embeddings.
To simulate the verification scenario, we compute pairwise similarity scores between each centroid embedding in the verification set and every centroid embedding in the enrollment set. Cosine similarity is used as the similarity metric, quantifying the angular closeness between embeddings in the feature space. A verification attempt is labeled as \textit{genuine} if the compared enrollment and verification embeddings are derived from recordings of the same participant; otherwise, the attempt is designated as an \textit{impostor}.

\subsection{Performance Metrics}

The resulting similarity scores and their corresponding labels are used to compute a Receiver Operating Characteristic (ROC) curve. From this curve, we derive several performance evaluation metrics:

\begin{itemize}

\item \textbf{Equal Error Rate (EER)}
\begin{itemize}
\item EER is defined as the point on the ROC curve where the False Rejection Rate (FRR) equals the False Acceptance Rate (FAR).
\item A lower EER indicates better overall authentication performance.
\item If an exact intersection point does not exist, EER is estimated using linear interpolation.
\item Computation of EER requires disjoint enrollment and authentication datasets.
\end{itemize}

\item \textbf{FRR at a Fixed FAR}
\begin{itemize}
\item FRR is reported at a predefined FAR threshold and denoted as $\text{FRR}_{X\%}$, where $X\%$ specifies the FAR. 
Though FRR at a 1-in-10,000 FAR is common among prior research, we have used FRR
at a 1-in-50,000 FAR to compare with the prior study \cite{lohr2024baseline} conducted on the same dataset.
\item For instance, $\text{FRR}_{0.002\%} = 1\%$ implies that the system falsely rejects genuine users 1\% of the time while maintaining a FAR of 0.002\% (equivalent to 1 in 50,000).
\item This metric assesses practical usability under stringent security constraints, aligning with industry standards such as FIDO~\cite{FIDO2020}.
\end{itemize}

\end{itemize}

\section{EXPERIMENTAL DESIGN}

Our experiments are designed to systematically investigate how key factors influence the authentication performance of our methodology. 
To this end, we conduct rigorous evaluations by varying the following factors:

\subsection{Authentication Scenarios ( $\times$ 2)}

Each participant contributed two independent sets of calibration parameters.
To examine how calibration affects verification performance, we tested the model under three distinct scenarios, which differ solely in the choice of calibration parameters used to estimate the visual axis during verification.
The scenarios are defined as follows and are illustrated in the diagram below:

\begin{figure*}[ht!]
    \centering        
    \includegraphics[width=0.75\columnwidth]{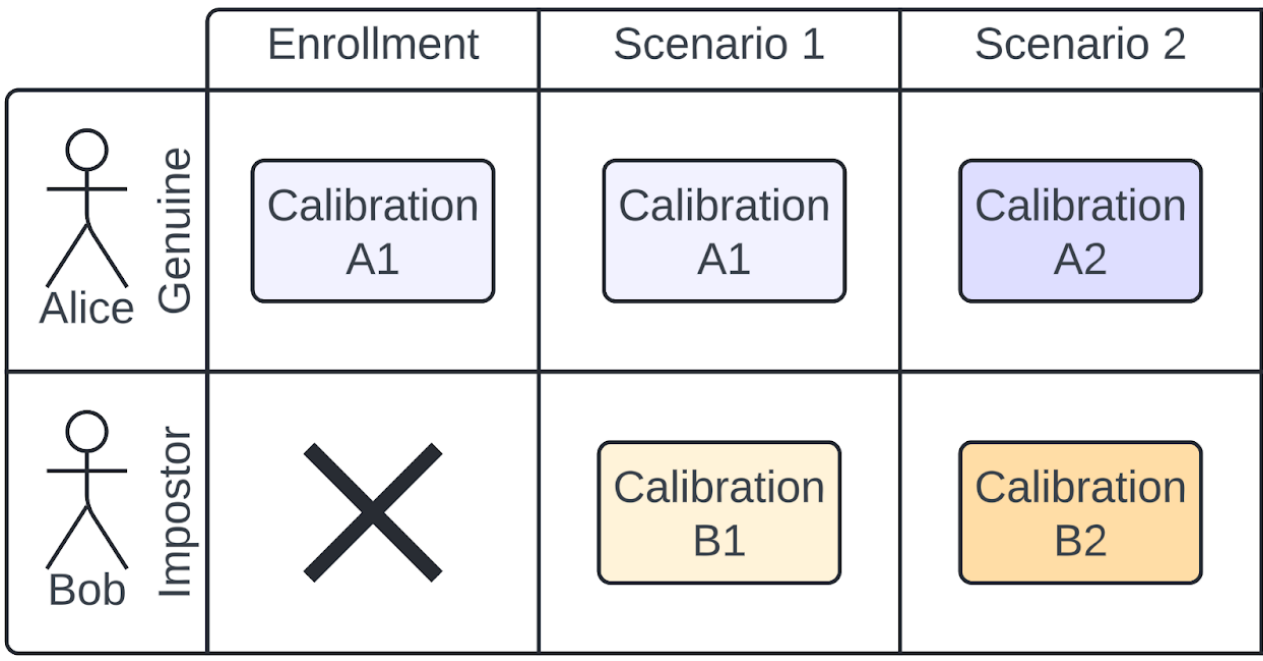}
    \caption{Diagram of which set of calibration parameters is used to estimate visual axis during enrollment and each verification scenario for both genuine and impostor verification attempts.  Alice is enrolled in the system.  At verification time, both Alice and an impostor Bob claim to be Alice.}
    \label{fig:scenarios}
\end{figure*}

\begin{enumerate}
\item \textbf{Scenario 1 ---} 
The verifying user’s first calibration is used to estimate the visual axis during verification.
For \textit{genuine} verification attempts, this calibration matches the one used during enrollment.
For \textit{impostor} attempts, the impostor’s calibration differs from the enrolled user’s calibration.
This is a scenario where an authentic and an impostor user must go through ET calibration procedure to try to gain access to the system.

\item \textbf{Scenario 2 ---}
The verifying user’s second ET calibration at different depth is used during verification. 
For all matches, this calibration is always different from the one used for enrollment, introducing a deliberate calibration mismatch even for genuine users. 
This scenario represents the case when a gaze-aided XR system requires different ET calibration depths for its operation.
Calibrating at one depth and using eye tracking data calibrated at a different depth was never used for authentication purposes before, which represents the novelty of this paper.

\end{enumerate}


\subsection{Calibration Parameters during Training ( $\times$ 2)}
Each participant provides two independent sets of calibration parameters, obtained from two separate random-jumping-dot tasks with gaze targets presented at different depths (200~cm and 75~cm). 
Those calibration parameters are used to convert the optical axis to the visual axis.

We investigate two approaches for utilizing these calibration sets during training:
\begin{itemize}
\item \textbf{Single}: Only one set of calibration parameters is applied to compute the visual axis estimates for each recording in the training set.
\item \textbf{All}: Both available calibration sets are used, generating two possible visual axis estimates per recording—one for each calibration set.
\end{itemize}

\subsection{ Signal Quality ( $\times$ 2)}






We compare authentication performance using gaze signals produced by two iterations of our end-to-end, machine–learning–based gaze estimation pipeline, both drawing conceptual inspiration from the methodology underlying Meta’s Project Aria Eye Tracking approach \cite{aria}. The \emph{New} pipeline incorporates multiple internal improvements relative to the \emph{Old} version; however, the specific implementation details are proprietary and therefore outside the scope of this paper.

What matters for our analysis is that the \emph{New} pipeline delivers measurably higher signal quality. 
For this reason, rather than discussing architectural or algorithmic differences, we distinguish the two pipelines directly in terms of their resulting signal quality profiles. 
Our focus is on how these differences—primarily reflected in spatial accuracy and spatial precision—affect authentication performance.
Follow Table \ref{tab:binocular_spatial} to compare the improvements by signal quality.
This framing allows us to evaluate how the intrinsic quality of the gaze signal influences downstream authentication outcomes.

\begin{table}[h!]
\centering
\caption{Binocular spatial accuracy (excluding fixations with dispersion $\geq$ 10 degrees) and spatial precision (S2S RMS) for \emph{New} and \emph{Old} Pipelines at 75 cm and 200 cm target depth. 
Metrics reported as median of medians.}
\begin{tabular}{lcc}
\hline
\textbf{Condition} & \textbf{Accuracy (deg)} & \textbf{Precision (deg)} \\
\hline
New Pipeline @ 75 cm & 0.79 & 0.20 \\
New Pipeline @ 200 cm & 0.79 & 0.20 \\
\hline
Old Pipeline @ 75 cm & 1.07 & 0.32 \\
Old Pipeline @ 200 cm & 1.07 & 0.32 \\
\hline
\end{tabular}
\label{tab:binocular_spatial}
\end{table}

\subsection{Axis of operation ( $\times$ 3)}

We evaluate the authentication performance of our approach using gaze estimates derived from different axes of operation: the optical axis (O), the visual axis (V), and a combination of both (B). This allows us to assess how different gaze estimation influences authentication performance.

It is important to restate that the appearance-based gaze estimation method used for the device that captured the data was similar to Meta's Project Aria~\cite{aria} eye-tracking approach. 
By optical axis, we refer to the raw, uncalibrated gaze direction estimated directly from the eye images, representing the geometric orientation of the eyeball without any personal calibration. 
A personalized linear calibration is then applied to map this optical axis to the true gaze vector (visual axis), thereby accounting for individual differences such as angle kappa. 
In short, the optical axis in our study represents an uncalibrated gaze estimate, and when combined with per-user calibration, it yields an accurate approximation of the visual axis.

\subsection{Epochs and Minibatch Size ($\times$ 2)}

We evaluate the model performance under two different training regimes, varying the number of training epochs and the minibatch size ($m$):

\begin{enumerate}
\item \textbf{Configuration 1}: Training for 100 epochs with $m$ = 256 (16 $\times$ 16).

\item \textbf{Configuration 2}: Training for 1000 epochs with $m$ = 1024 (32$\times$ 32).
\end{enumerate}

\subsection{Filtering ($\times$ 2)}

We examine the impact of simple temporal smoothing on the authentication performance by introducing a lightweight filtering factor with two configurations:

\begin{itemize}
\item \textbf{Filter On}:
A simple 3-sample moving average filter is applied to the raw gaze position signals before passing them into the further data preprocessing step.
This filter smooths short-term noise and small tracking fluctuations, which can otherwise introduce variability in estimated eye movement velocities. The filter operates with a window size of three consecutive samples, updating sequentially without future information.

\item \textbf{Filter Off}:
No temporal smoothing is applied. 

\end{itemize}

\subsection{Statistical Test}
The analysis employs a pairwise comparison strategy to isolate the effect of each experimental factor on the system's performance metrics (EER/FRR). 
Table \ref{tab:stat_param} summarizes the key components used in the statistical tests.

The T-statistic is calculated based on the difference between the two sample means, scaled by the estimated standard error of that difference. Assuming the variances are equal (a reasonable assumption for this type of experimental data), the T-statistic is calculated as:

\begin{equation}
t = \frac{\bar{x}_1 - \bar{x}_2}{s_p \cdot \sqrt{\frac{1}{N_1} + \frac{1}{N_2}}}
\end{equation}

Where:
\begin{itemize}
    \item $\bar{x}_1$ and $\bar{x}_2$ are the sample means of the two conditions.
    \item $N_1$ and $N_2$ are the sample sizes (both equal to 4).
    \item $s_p$ is the pooled standard deviation, which is calculated as:
\end{itemize}

\begin{equation}
    s_p = \sqrt{\frac{(N_1-1)s_1^2 + (N_2-1)s_2^2}{N_1 + N_2 - 2}}
\end{equation}

The calculated P-value from the T-statistic is then used to determine if the Null Hypothesis ($H_0$) should be rejected.

\begin{table}[htbp]
\centering
\caption{Key Statistical Parameters Used in Comparison}
\label{tab:key_parameters}
\begin{tabularx}{\columnwidth}{l X}
\toprule
\textbf{Component} & \textbf{Value/Description} \\
\midrule
\textbf{Metric} & $\text{EER}$ and $\text{FRR at a Fixed FAR}$. \\
\textbf{Sample Size} $(N)$ & $N=4$ for all experiments. This represents the number of independent metric measurements derived from the four data folds. \\
\textbf{Statistical Test} & Two-sample t-test for independent means (unpaired t-test). \\
\textbf{Significance Level} $(\alpha)$ & $\alpha = 0.05$. Results with P-value $< 0.05$ are considered statistically significant. \\
\textbf{Null Hypothesis} $(H_0)$ & There is no difference in the true mean biometric performance between the two experiments. \\
\bottomrule
\end{tabularx}
\label{tab:stat_param}
\end{table}

\section{RESULTS}

We summarize the authentication performance under various experimental conditions in Table~\ref{table:full_table}. 
Each experiment is assigned a unique ID to facilitate easy reference and comparison across conditions.

\begin{table*}[htbp]
\centering
\caption{Performance metrics across different experiments. 
Lower metric values indicate better performance. SD represents Standard deviation.
In the ``Experiment'' column, ``S1'' represents Scenario-1 and ``S2'' represents Scenario-2.
In the ``Axis'' column, ``O'' is the optical axis, “V” is the visual axis, and “B” is both.
`` · '' is used to indicate that the value is repeated from the row above.
The best-performing results are shown in bold.}
\begin{tabular}{cccccccc}
\hline
Experiment & Calibration & Pipeline & Axis & Epochs \& Minibatch & Filter & Mean EER(\%) (SD) &  Mean FRR$_{0.002\%}$ (\%) (SD) \\ \hline
S1@1    & All & New   & O & 100 \& $m$ = 256  & Off  & 5.75 (0.13)& 87.73 (0.83)\\  
S1@2    & ·   & ·   & V & ·  & ·  & 4.51 (0.61)& 81.01 (3.79)\\
S1@3    & ·   & ·   & B & · & · & 0.30 (0.03)& 43.08 (6.78)\\  
S1@4    & ·   & ·   & B & 1000 \& $m$ = 1024 & · & 0.11 (0.04)& 10.37 (2.49)\\
S1@5    & ·   & ·   & V & 100 \& $m$ = 256 & On& 5.45 (0.16)& 87.11 (0.79)\\ 
\hline

S1@6   & All   & Old   & O & 100 \& $m$ = 256  & Off  & 7.64 (0.20)& 92.88 (0.53)\\  
S1@7   & ·   & ·   & V &  · & · & 5.05 (0.25) & 86.14 (0.90)\\  
S1@8   & ·   & ·   & B & · & · & 0.17 (0.06) & 24.26 (5.93)\\  
S1@9   & ·   & ·   & B & 1000 \& $m$ = 1024 & · & 0.07 (0.02)& 6.08 (0.84)\\
S1@10   & ·   & ·   & V & 100 \& $m$ = 256 & On & 8.19 (0.13)& 94.70 (0.36)\\ 
\hline

S1@11    & Single    & New   & O & 100 \& $m$ = 256 & Off & 5.74 (0.23)& 88.34 (1.08)\\  
S1@12    & ·    & ·   & V & ·  & · & 2.79 (0.98) & 79.26 (4.31)\\  
S1@13    & ·    & ·   & B & · & · & 0.39 (0.08) & 51.31 (5.72)\\  
S1@14    & ·    & ·   & B & 1000 \& $m$ = 1024 & · & 0.07 (0.03)& 10.42 (7.86)\\
S1@15   & ·   & ·   & V & 100 \& $m$ = 256 & On& 5.61 (0.16)& 88.19 (0.65)\\ 
\hline

S1@16   & Single    & Old   & O & 100 \& $m$ = 256  & Off  & 7.68 (0.16) & 93.24 (0.48)\\  
S1@17   & ·    & ·   & V & ·  & ·  & 3.28 (0.17) & 97.03 (0.97)\\  
S1@18   & ·    & ·   & B & · & · & 0.18 (0.07) & 32.25 (15.22) \\  
S1@19   & ·    & ·   & B & 1000 \& $m$ = 1024 & · & \textbf{0.01 (0.01)} & \textbf{0.87 (0.23)} \\
S1@20   & ·    & ·   & V & 100 \& $m$ = 256 & On & 8.21 (0.19) & 94.83 (0.51)\\ 
\hline

&   &   &  & & & & \\ \hline

S2@1  & All    & New & O & 100 \& $m$ = 256 & Off & 5.76 (0.13) & 87.81 (0.84) \\ 
S2@2  & ·    & · & V & · & · & 6.04 (0.27) & 87.64 (0.84) \\
S2@3  & ·    & · & B & · & · & 5.80 (0.14) & 91.84 (1.22) \\ 
S2@4  & ·    & · & B & 1000 \& $m$ = 1024 & · & 6.09 (0.08) & 92.40 (0.83) \\
S2@5  & ·    & · & V & 100 \& $m$ = 256 & On & 5.73 (0.17) & 87.75 (0.67) \\ 
\hline

S2@6 & All    & Old & O & 100 \& $m$ = 256 & Off & 7.63 (0.22) & 92.94 (0.54) \\   
S2@7 & ·    & · & V & · & · & 11.12 (0.37) & 95.37 (0.37) \\   
S2@8 & ·    & · & B & · & · & 9.64 (0.23) & 96.39 (0.48) \\
S2@9 & ·    & · & B & 1000 \& $m$ = 1024 & · & 11.73 (0.34) & 97.40 (0.31) \\
S2@10 & ·    & · & V & 100 \& $m$ = 256 & On & 8.45 (0.17) & 94.98 (0.38) \\ 
\hline

S2@11  & Single & New & O & 100 \& $m$ = 256 & Off & 5.75 (0.22) & 88.40 (1.07) \\ 
S2@12  & · & · & V & · & · & 28.25 (2.08) & 99.83 (0.17) \\
S2@13  & · & · & B & · & . & 24.68 (1.77) & 98.72 (0.21) \\ 
S2@14  & · & · & B & 1000 \& $m$ = 1024 & · & 23.70 (1.47) & 98.79 (0.44) \\
S2@15  & · & · & V & 100 \& $m$ = 256 & On & 5.95 (0.17) & 88.89 (0.68) \\ 
\hline

S2@16 & Single & Old & O & 100 \& $m$ = 256 & Off & 7.66 (0.15) & 93.29 (0.49) \\   
S2@17 & · & · & V & · & · & 33.68 (1.23) & 99.99 (0.03) \\   
S2@18 & · & · & B & · & · & 35.15 (3.83) & 99.38 (0.28) \\
S2@19 & · & · & B & 1000 \& $m$ = 1024 & · & 36.63 (0.49) & 99.54 (0.09) \\ 
S2@20 & · & · & V & 100 \& $m$ = 256 & On & 8.52 (0.17) & 95.28 (0.49) \\ 
\hline

\end{tabular}
\label{table:full_table}
\end{table*}

\begin{table*}[htbp]
\centering
\caption{
Effect of the \emph{Authentication Scenario} (S1→S2) on EER and FRR$_{0.002\%}$\%, including statistical significance.
``S1'' and ``S2'' denote Scenario~1 and Scenario~2, respectively. 
In the ``Axis'' column, ``O'' indicates the optical axis, ``V'' the visual axis, and ``B'' both axes.
Change ($\Delta$) in biometric performance with $\text{P-value} < 0.05$ is considered statistically significant and added in the table, with the corresponding t-statistic appearing in parentheses. 
$\textcolor{red}{\downarrow}$ represents performance degrades whereas $\textcolor{green}{\uparrow}$ represents improvement in performance.
A ``---'' indicates non-significant results.}
\begin{tabular}{lccc}
\hline
\textbf{Compared Experiments} & \textbf{Axis} & $\Delta$ \textbf{EER\%} & $\Delta$ \textbf{FRR$_{0.002\%}$\%} \\
\hline
S1@1 vs.\ S2@1 & O & --- & --- \\ 
S1@6 vs.\ S2@6 & O & --- & --- \\ 
S1@11 vs.\ S2@11 & O & --- & --- \\
S1@16 vs.\ S2@16 & O & --- & --- \\
\hline
S1@2 vs.\ S2@2 & V & $\textcolor{red}{\downarrow}\ 1.53$ \, $(4.59)$ & $\textcolor{red}{\downarrow}\ 6.63$ \, $(3.41)$ \\
S1@7 vs.\ S2@7 & V & $\textcolor{red}{\downarrow}\ 6.07$ \, $(27.2)$ & $\textcolor{red}{\downarrow}\ 9.23$ \, $(18.98)$ \\
S1@12 vs.\ S2@12 & V & $\textcolor{red}{\downarrow}\ 25.46$ \, $(22.16)$ & $\textcolor{red}{\downarrow}\ 20.57$ \, $(9.40)$ \\
S1@17 vs.\ S2@17 & V & $\textcolor{red}{\downarrow}\ 30.40$ \, $(48.98)$ & $\textcolor{red}{\downarrow}\ 2.96$ \, $(6.10)$ \\
\hline
S1@3 vs.\ S2@3 & B & $\textcolor{red}{\downarrow}\ 5.50$ \, $(76.8)$ & $\textcolor{red}{\downarrow}\ 48.76$ \, $(14.15)$ \\ 
S1@8 vs.\ S2@8 & B & $\textcolor{red}{\downarrow}\ 9.47$ \, $(79.6)$ & $\textcolor{red}{\downarrow}\ 72.13$ \, $(24.25)$ \\
S1@13 vs.\ S2@13 & B & $\textcolor{red}{\downarrow}\ 24.29$ \, $(27.4)$ & $\textcolor{red}{\downarrow}\ 47.41$ \, $(16.58)$ \\
S1@18 vs.\ S2@18 & B & $\textcolor{red}{\downarrow}\ 34.97$ \, $(18.26)$ & $\textcolor{red}{\downarrow}\ 67.13$ \, $(8.83)$ \\
\hline
\end{tabular}
\label{tab:s1s2_differences}
\end{table*}



\textit{Baseline comparison--- } Our baseline is Experiment S1@1 corresponds to Scenario~1 in the authentication scenarios. 
This baseline configuration uses the new pipeline (enhanced signal quality), all calibration parameters during training, visual‐axis gaze estimation, and filter off.
Unless otherwise specified, results were computed using data from the random saccade (jumping dot) task, with the full training set (6,134 recordings) and the full test set (1,983 users). 
The model was trained for 100 epochs with a minibatch size of $m = 256$ samples, and both enrollment and verification used 20 seconds of gaze data.
EER and FRR$_{0.002\%}$ are presented as ``mean (SD)'' across a 10-fold split of the dataset.
For comparison, Lohr et al.~\cite{lohr2025ocular} used both eyes’ optical axes for eye movement–based authentication and reported an EER of 5.75\% and a FRR$_{0.002\%}$ of 87.73\%. 
In our baseline experiment S1@1, we adopt the same principle and obtain identical results. 

\textit{Best performance comparison--- } We achieved our best performance from Experiment S1@19 with an EER of 0.01\% and an FRR$_{0.002\%}$ of 0.87\%. 
This is comparable to Experiment-6 from Lohr et al. \cite{lohr2024baseline}, which reported an EER of 0.04\% and an FRR$_{0.002\%}$ of 2.4\% under the same setting. Overall, our best case represents a substantial improvement over the current state-of-the-art.





Following, we will examine how key factors influence the authentication performance of our methodology. 
Specifically, we aim to identify which factors contribute most to performance variations and understand their underlying impact on the reliability and robustness of the authentication system.

\subsection{Authentication Scenarios}
\label{sec:s1s2_differences}

Table \ref{tab:s1s2_differences} presents the statistical significance of performance differences between authentication scenarios (S1 vs S2) across multiple experimental conditions. 
Each comparison corresponds to identical experimental setups in both scenarios, as defined in Table \ref{table:full_table}. 
The results are categorized based on the axis.

This trend is the most pronounced among all factors considered. 
As calibration is used solely to estimate the visual axis, S2 does not affect the optical axis.
Therefore, as expected, there is no statistically significant performance difference between scenarios for the optical axis.

In contrast, when the visual axis or both axes were involved, a consistent significant degradation in performance was observed.
For experiments involving only the visual axis with filter off, EER increases ranged from +1.53\% to +30.40\%, while FRR$_{0.002\%}$ increases ranged from +2.96\% to +20.57\%.
Experiments involving both axes with filter off, EER degradation was more pronounced, ranging from +5.50\% to +34.97\%, with corresponding FRR$_{0.002\%}$ increases reaching up to +72.13\%.

\subsection{Calibration Parameters during Training}
\label{sec:callib}

Table \ref{tab:calibration_effect} presents a significant difference in performance that varied because of the calibration parameters used during training in S1.
As discussed in Section \ref{sec:s1s2_differences}, performance degrades in S2 (eye movement data is calibrated via calibration procedure at different depths) regardless of the calibration approach. 
Therefore, Table \ref{tab:calibration_effect} focuses solely on S1. 

The results indicate that the choice between \textbf{All} and \textbf{Single} calibration has a limited influence on performance in S1, with only minor variations observed.
The comparison between S1@2 and S1@12 revealed a slight improvement in EER only.
EER improves for S1@7 and S1@17 comparison as well, but  FRR$_{0.002\%}$ degraded noticeably (by 10.89\%), indicating a mixed benefit of \textbf{Single} calibration approach in visual axis.
In short, other than visual-axis \& filter On pair, there is no significant difference in performance.
Overall, the findings indicate that the choice of calibration approach—whether \textbf{All} or \textbf{Single}—does not substantially influence biometric performance in S1.

\begin{table}[htbp]
\centering
\caption{
Effect of the \emph{Calibration Parameters} (All→Single) on EER and FRR$_{0.002\%}$\%, including statistical significance.
Other details as in Table \ref{tab:s1s2_differences}.}
\begin{tabular}{lccc}
\hline
\textbf{Compared Experiments} & \textbf{Axis} & $\Delta$ \textbf{EER\%} & $\Delta$ \textbf{FRR$_{0.002\%}$\%} \\
\hline
S1@1 vs.\ S1@11 & O & --- & --- \\
S1@2 vs.\ S1@12 & V & $\textcolor{green}{\uparrow}\ 1.72$ \, (2.98) & --- \\
S1@3 vs.\ S1@13 & B & --- & --- \\
S1@5 vs.\ S1@15 & V & --- & --- \\
S1@6 vs.\ S1@16 & O & --- & --- \\
S1@7 vs.\ S1@17 & V & $\textcolor{green}{\uparrow}\ 1.77$ \, (11.7) & $\textcolor{red}{\downarrow}\ 10.89$ \, (16.4) \\
S1@8 vs.\ S1@18 & B & --- & --- \\
S1@10 vs.\ S1@20 & V & --- & --- \\
\hline
\end{tabular}
\label{tab:calibration_effect}
\end{table}

\subsection{Signal Quality}
\label{sec-sigQ}

As discussed in Section \ref{sec:s1s2_differences}, performance degrades in S2 regardless of the signal quality (different gaze-estimation pipeline). 
Therefore, this analysis focuses solely on S1.
This section examines the impact of signal quality on authentication performance by comparing the \emph{New} pipeline with the \emph{Old} pipeline within S1.
Table \ref{tab:signal_quality_s1} summarizes the statistical significance of changes in EER and FRR$_{0.002\%}$ across identical experimental setups.

\begin{table}[htbp]
\centering
\caption{Effect of the \emph{Signal Quality} (New→Old) on EER and FRR$_{0.002\%}$\%, including statistical significance.
Other details as in Table \ref{tab:s1s2_differences}.}
\begin{tabular}{lccc}
\hline
\textbf{Compared Experiments} & \textbf{Axis} & $\Delta$ \textbf{EER\%} & $\Delta$ \textbf{FRR$_{0.002\%}$\%} \\
\hline
S1@1  vs.\ S1@6  & O & $\textcolor{red}{\downarrow}\ 1.89$ \, (15.84) & $\textcolor{red}{\downarrow}\ 5.15$ \, (10.47) \\
S1@2  vs.\ S1@7  & V & --- & $\textcolor{red}{\downarrow}\ 5.13$ \, (2.63) \\
S1@3  vs.\ S1@8  & B & $\textcolor{green}{\uparrow}\ 0.13$ \, (3.88) & $\textcolor{green}{\uparrow}\ 18.82$ \, (4.18) \\
S1@5  vs.\ S1@10 & V & $\textcolor{red}{\downarrow}\ 2.74$ \, (26.6) & $\textcolor{red}{\downarrow}\ 7.59$ \, (17.45) \\
\hline
S1@11 vs.\ S1@16 & O & $\textcolor{red}{\downarrow}\ 1.94$ \, (13.8) & $\textcolor{red}{\downarrow}\ 4.90$ \, (8.29) \\
S1@12 vs.\ S1@17 & V & --- & $\textcolor{red}{\downarrow}\ 17.77$ \, (8.05) \\
S1@13 vs.\ S1@18 & B & $\textcolor{green}{\uparrow}\ 0.21$ \, (3.95) & --- \\
S1@15 vs.\ S1@20 & V & $\textcolor{red}{\downarrow}\ 2.60$ \, (20.94) & $\textcolor{red}{\downarrow}\ 6.64$ \, (16.07) \\
\hline
\end{tabular}
\label{tab:signal_quality_s1}
\end{table}

In most of the experiments involving the optical and visual axes, the \emph{New} pipeline resulted in better performance. 
For example, S1@1 → S1@6 showed there is statistically significant improved performance in both EER and FRR$_{0.002\%}$ with the \emph{New} pipeline. 
Similarly, S1@5 → S1@10, S1@11 → S1@16 and S1@15 → S1@20 showed improved performance with the \emph{New} pipeline.
Except for some cases, all cases involving visual axis (whether filter on or off), \emph{New} pipeline yields better performance.

Interestingly, experiments involving both axes showed better performance with the \emph{Old} pipeline. 
For instance, S1@3 → S1@8 and S1@13 → S1@18 showed small but significant decreases in EER and substantial reductions in FRR$_{0.002\%}$ .

So,  the \emph{New} pipeline with better signal quality yields better performance individually for optical and visual axes. But for both axis set-up \emph{Old} is still yielding better performance.


\subsection{Optical vs Visual vs Both axis}

Table \ref{tab:axis} compares the performance relation between the optical, the visual axis, and both axes.

In S1, both axis always outperforms the individual axis, and the visual axis mostly yields better performance than optical, with only one exception.
For example (referencing to Table \ref{table:full_table}), the optical-only baseline (S1@1) achieved an EER of 5.75\% and a FRR$_{0.002\%}$ of 87.73\%. 
Switching to the visual axis (S1@2) reduced the EER to 4.51\% (FRR$_{0.002\%}$ of 81.01\%).
Feeding both axes into the model (S1@3) yielded a further improvement, achieving an EER of 0.30\% and an FRR$_{0.002\%}$ of 43.08\%.

\begin{table}[htbp]
\centering
\caption{Axis-wise ordering across three-step progressions for S1 and S2.
EER and FRR$_{0.002\%}$ are compared across axes.
A ``$>$'' indicates a statistically significant difference; ``$\equiv$'' indicates a non-significant difference.}
\label{tab:axis}
\begin{tabularx}{0.57\columnwidth}{lcc}
\hline
\textbf{Compared Experiments} &  $\Delta$ \textbf{EER\%} & $\Delta$ \textbf{FRR$_{0.002\%}$\%} \\
\hline
S1@1  $\rightarrow$ S1@2  $\rightarrow$ S1@3     & B $>$ V $>$ O   & B $>$ V $>$ O \\
S1@6  $\rightarrow$ S1@7  $\rightarrow$ S1@8     & B $>$ V $>$ O   & B $>$ V $>$ O \\
S1@11 $\rightarrow$ S1@12 $\rightarrow$ S1@13    & B $>$ V $>$ O   & B $>$ V $>$ O \\
S1@16 $\rightarrow$ S1@17 $\rightarrow$ S1@18    & B $>$ V $>$ O   & B $>$ O $>$ V \\
\hline
S2@1  $\rightarrow$ S2@2  $\rightarrow$ S2@3     & O $\equiv$ B $\equiv$ V & V $\equiv$ O $>$ B \\
S2@6  $\rightarrow$ S2@7  $\rightarrow$ S2@8     & O $>$ B $>$ V   & O $>$ V $>$ B \\
S2@11 $\rightarrow$ S2@12 $\rightarrow$ S2@13    & O $>$ B $>$ V   & O $>$ B $>$ V \\
S2@16 $\rightarrow$ S2@17 $\rightarrow$ S2@18    & O $>$ V $\equiv$ B & O $>$ B $>$ V \\
\hline
\end{tabularx}

\end{table}

In S2, results were less consistent, with varied orderings across experiments. 
However, as discussed in Section \ref{sec:s1s2_differences}, S2 does not affect the optical axis, leading to the optical axis often being the best performer in S2.




In summary, both axes consistently provide the best performance in S1, followed by the visual axis, with the optical axis performing the worst. 
In S2, performance rankings are less stable, with the optical axis generally yielding the best performance due smaller calibration depth effect on GA performance.


\subsection{Training Depth}
\label{sec:training}

Increasing epochs and minibatch size didn't affect performance for optical or visual axes individually. 
Hence, these results are not included in Table \ref{table:full_table}.
However, as shown in Table \ref{tab:epoch_change_analysis}, increasing the number of training epochs from 100 to 1000 and the minibatch size from $m$ = 256 to $m$ = 1024 yielded substantial improvements in performance for experiments involving the visual axis combined with the optical axis in Scenario-1.
For example, S1@3 vs. S1@4 and S1@18 vs. S1@19 showed significant improvement is achieved in both EER and FRR$_{0.002\%}$ with longer training depth. 
The overall best performance is also observed with increased training epochs and minibatch size in S1@19 (EER of 0.01 and FRR$_{0.002\%}$ of 0.87).

In contrast, in S2, increasing training depth resulted in a mixed effect, with most comparisons showing either negligible changes or slight performance degradation. 
Given that performance in S2 is generally lower across experiments, these results suggest that the observed degradation is more likely due to S2 conditions rather than the effect of training depth itself.

\begin{table}[htbp]
\centering
\caption{Effect of the \emph{Training Depth} (epoch 100 \& m = 256 → epoch 1000 \& m = 1024) on EER and FRR$_{0.002\%}$\%, including statistical significance.
Other details as in Table \ref{tab:s1s2_differences}. Both axis only.}
\begin{tabular}{lcc}
\hline
\textbf{Compared Experiments} & $\Delta$ \textbf{EER\%} & $\Delta$ \textbf{FRR$_{0.002\%}$\%} \\
\hline
S1@3  vs.\ S1@4  & $\textcolor{green}{\uparrow}\ 0.19$ \, (7.60) & $\textcolor{green}{\uparrow}\ 32.71$ \, (9.06) \\
S1@8  vs.\ S1@9  & $\textcolor{green}{\uparrow}\ 0.10$ \, (3.16) & $\textcolor{green}{\uparrow}\ 18.18$ \, (6.07) \\
S1@13 vs.\ S1@14 & $\textcolor{green}{\uparrow}\ 0.32$ \, (7.49) & $\textcolor{green}{\uparrow}\ 40.89$ \, (8.41) \\
S1@18 vs.\ S1@19 & $\textcolor{green}{\uparrow}\ 0.17$ \, (4.81) & $\textcolor{green}{\uparrow}\ 31.38$ \, (4.12) \\
\hline
S2@3  vs.\ S2@4  & $\textcolor{red}{\downarrow}\ 0.29$ \, (3.60) & --- \\
S2@8  vs.\ S2@9  & $\textcolor{red}{\downarrow}\ 2.09$ \, (10.18) & $\textcolor{red}{\downarrow}\ 1.01$ \, (3.54) \\
S2@13 vs.\ S2@14 & --- & --- \\
S2@18 vs.\ S2@19 & --- & --- \\
\hline
\end{tabular}
\label{tab:epoch_change_analysis}
\end{table}


\subsection{Filtering}
\label{sec:filtering}

In this analysis, we did not have access to filtered optical axis measurements. 
Consequently, conditions involving filtered optical axis or both axes are omitted from the results, and only visual axis-filtered data is included in Table \ref{tab:fiter_change}.

For S1, the t-tests indicate statistically significant degradation in nearly all comparisons. 
EER and FRR$_{0.002\%}$ increase after filtering in every transition except S1@17→S1@20, where broader performance losses overshadow the small FRR improvement. 
This indicates that filtering removes information that S1 relies on for discrimination.

In contrast, enabling the filter in S2 generally improved performance: S2@12 → S2@15 showed a substantial decrease in EER (22.30\%) and FRR$_{0.002\%}$ (10.94\%), S2@17 → S2@20 similarly showed strong improvements with EER decreasing by 25.16\% and FRR$_{0.002\%}$ by 4.71\%.
Most of the performance improvement is statistically significant.
Enabling filtering is the only case where performance improves in S2.

\begin{table}[htbp]
\centering
\caption{Effect of \emph{Filtering} on EER and FRR$_{0.002\%}$\%, including statistical significance.
Other details as in Table \ref{tab:s1s2_differences}. Visual axis only.}
\begin{tabular}{lccc}
\hline
\textbf{Compared Experiments}  & $\Delta$ \textbf{EER\%} & $\Delta$ \textbf{FRR$_{0.002\%}$\%} \\
\hline
S1@2  vs.\ S1@5  & $\textcolor{red}{\downarrow}\ 0.94$ \, (2.98) & $\textcolor{red}{\downarrow}\ 6.10$ \, (3.15) \\
S1@7  vs.\ S1@10 & $\textcolor{red}{\downarrow}\ 3.14$ \, (22.29) & $\textcolor{red}{\downarrow}\ 8.56$ \, (17.66) \\
S1@12 vs.\ S1@15 & $\textcolor{red}{\downarrow}\ 3.82$ \, (5.68)  & $\textcolor{red}{\downarrow}\ 8.93$ \, (4.10) \\
S1@17 vs.\ S1@20 & $\textcolor{red}{\downarrow}\ 4.93$ \, (38.67) & $\textcolor{green}{\uparrow}\ 2.20$ \, (4.01) \\
\hline
S2@2  vs.\ S2@5  & --- & --- \\
S2@7  vs.\ S2@10 & $\textcolor{green}{\uparrow}\ 2.67$ \, (13.11) & --- \\
S2@12 vs.\ S2@15 & $\textcolor{green}{\uparrow}\ 22.30$ \, (21.37) & $\textcolor{green}{\uparrow}\ 10.94$ \, (31.22) \\
S2@17 vs.\ S2@20 & $\textcolor{green}{\uparrow}\ 25.16$ \, (40.53) & $\textcolor{green}{\uparrow}\ 4.71$ \, (19.19) \\
\hline
\end{tabular}
\label{tab:fiter_change}
\end{table}

As Table \ref{tab:fiter_change} shows a consistent divergence in how filtering affects performance across the two calibration-depth settings. 
In S1, filtering reduces authentication performance, whereas in S2 it generally improves it. 
Because calibration depth is the only factor that differs between S1 and S2, this contrast points to an interaction between filtering and the depth-related characteristics of the gaze signal.
We have conducted another comparison.

Referring to Table \ref{tab:fiter_change_dis}, when filtering is Off in both, the authentication performance drop from S1 to S2 is substantial and statistically significant. 
On the other hand, when filtering is On in S2, the drop is still present and significant but considerably smaller. 
In other words, when calibration depth varies, filtering appears to reduce some of the added instability introduced by the depth change. 
This reversal relative to S1 suggests that filtering may be compensating for depth-related noise in S2, even though it suppresses informative detail under stable depth.

\begin{table}[htbp]
\centering

\caption{Change in EER and FRR$_{0.002\%}$ between S1 Filter Off, S2 Filter Off and On conditions (S1 Filter Off $\rightarrow$ S2 Filter Off \& S1 Filter Off $\rightarrow$ S2 Filter On). Visual axis only.}
\begin{tabular}{cccc}
\hline
\multicolumn{2}{c}{\textbf{Compared Experiments}} & $\Delta$ \textbf{EER (\%)} & $\Delta$\textbf{FRR$_{0.002\%}$} \\
\hline
\multirow{2}{*}{S1@2$\rightarrow$}       & S2@2       & $\textcolor{red}{\downarrow}\ 1.53$   & $\textcolor{red}{\downarrow}\ 6.63$ \\
& S2@5 & $\textcolor{red}{\downarrow}\ 1.22$   & $\textcolor{red}{\downarrow}\ 6.74$ \\
\hline
\multirow{2}{*}{S1@7$\rightarrow$}       & S2@7       & $\textcolor{red}{\downarrow}\ 6.07$   & $\textcolor{red}{\downarrow}\ 9.23$ \\
                            & S2@10 & $\textcolor{red}{\downarrow}\ 3.4$   & $\textcolor{red}{\downarrow}\ 8.84$ \\
\hline
\multirow{2}{*}{S1@12$\rightarrow$}      & S2@12 & $\textcolor{red}{\downarrow}\ 25.46$  & $\textcolor{red}{\downarrow}\ 20.57$ \\
                            & S2@15  &    $\textcolor{red}{\downarrow}\ 3.16$  & $\textcolor{red}{\downarrow}\ 9.63$ \\
\hline
\multirow{2}{*}{S1@17$\rightarrow$}      & S2@17 & $\textcolor{red}{\downarrow}\ 30.40$  & $\textcolor{red}{\downarrow}\ 2.96$ \\ 
                            & S2@20      & $\textcolor{red}{\downarrow}\ 5.24$  & $\textcolor{green}{\uparrow}\ 1.75$ \\ \hline
\end{tabular}
\label{tab:fiter_change_dis}
\end{table}

\section{DISCUSSION}
This study aimed to systematically investigate how key factors influence the authentication performance of our methodology, rather than to develop state-of-the-art benchmarks.


\subsection{RQ Analysis}
In the following, we discuss our findings concerning the research questions stated earlier.

\paragraph{(RQ1) How does calibration/re-calibration affect gaze authentication accuracy?}
Calibration (or recalibration) is required because the eye movement-based authentication system estimates the visual axis by mapping eye images to gaze points using user-specific calibration parameters.
Without calibration, the system can only access the optical axis.
It lacks the important individual gaze mapping, leading to degraded authentication accuracy. 
Recalibration becomes necessarily important whenever the user’s calibration parameters change due to environmental changes or device repositioning, or, as examined in our study, intentional changes in depth of calibration targets. 
As shown in Table \ref{tab:s1s2_differences}, recalibrating at different depths (Scenario 1 vs 2) consistently degrades authentication performance. 
This information should be taken into consideration when designing a GA system with an underlying appearance-based gaze estimation pipeline.

\paragraph{(RQ2) What's the effect of different calibration settings (during training) on performance?}
The analysis shows that using either a Single set or All sets of calibration parameters during training yields minimal differences in authentication performance for S1. 
In most comparisons, there is no substantial change in EER or FRR$_{0.002\%}$, indicating that the model’s ability to generalize across recordings is robust to the calibration parameters used during the training phase.
Though there are mixed cases where minor gains are observed, however, the choice between using all calibration sets or a single calibration set during training does not substantially affect the system’s performance in S1.
This shows that it is possible to simplify the training procedure for the GA-based system and for authentication purposes, use the same target depth for calibration, enrollment and verification.

\paragraph{(RQ3) How does ET signal quality, specifically spatial accuracy and precision, affect authentication accuracy?}
Our results (Table \ref{tab:signal_quality_s1}) demonstrate that signal quality significantly impacts authentication performance, but its effect is modulated by the specific axes used.
Overall, the \emph{New} pipeline, characterized by better spatial accuracy (median 0.79° vs. 1.07°) and spatial precision (median RMS 0.20° vs. 0.32°), yields improved authentication performance when either the optical or visual axis is used individually.
Interestingly, experiments using both axes as input to the authentication system showed better performance with the \emph{Old} pipeline.

\paragraph{(RQ4) How does the performance provided by the visual axis compare to the performance of the optical axis?} 
The results indicate that both the optical and visual axes together yield the best authentication performance, particularly under the S1 configuration. 
In S1, combining both axes consistently outperformed using either axis alone, with the visual axis generally performing better than the optical axis.

Under the S2 configuration, the optical axis often yields better results, not because its performance improves, but because the visual axis performance drops significantly.
This degradation appears to be driven by S2’s specific condition changes, which affect the visual axis but leave the optical axis largely unaffected. 
As a result, the optical axis emerges as the more robust reference under S2, despite its performance remaining relatively stable (compared to S1).
Therefore, we conclude that both axes should be employed for best GA performance for a system with the same-depth calibration condition, whereas the optical axis alone is preferable when calibration conditions vary, thus highlighting the importance of the ocular authentication system described here \cite{lohr2025ocular}.

\paragraph{(RQ5) What's the impact of training depth?}
The training depth has a huge impact on the authentication accuracy.
By extending training up to ten times longer and increasing the minibatch size by a factor of four, we achieved our best result to date (compared to Experiment 6 of \cite{lohr2024baseline}), reaching an FRR$_{0.002\%}$ of just 0.87\% in Experiment S1@19.
This outcome is particularly impressive because it shows that with only 20 seconds of gaze data sampled at 72 Hz, our system can satisfy the FIDO security requirements \cite{FIDO2020}, which specify a maximum FRR of 3\% at a false acceptance rate of 1 in 50,000, all within a 30-second verification window. 
It is important to keep in mind that these results are based on a short test-retest interval.
We would expect the performance to decline as the time between enrollment and verification increases, based on prior research \cite{lohr_ekyt}. 

In addition, the reliability and overall performance of the authentication system warrant consideration.
Findings from \cite{raju2024temporal} indicate that downsampling to as low as 100 Hz can still yield high reliability and strong authentication performance.
Our results are consistent with this observation, further supporting the robustness of the system at lower sampling rates.
Given that we meet the FIDO requirement using only 20 seconds of data (within the allowable 30-second verification window), further investigation into even lower sampling rates—below 72 Hz—would be a valuable direction for future work.

\paragraph{(RQ6) Do we need filtering to improve authentication accuracy?}

Prior work has shown that smoothing-based filters—such as the three-sample moving-average filter used here—tend to degrade authentication performance under the same calibration depth, largely because the filter suppresses identity-relevant detail while increasing the privacy of the signal \cite{raju2025real}.
The S1 results align with this established finding.
The behavior in S2 is different. 
Filtering appears to function as a stabilizing mechanism. 
When it is disabled, the shift from S1 to S2 produces a marked decline in authentication performance.
With filtering enabled in S2, the decline persists but is substantially attenuated. 
This pattern indicates that depth variation introduces additional noise, and filtering mitigates a portion of this instability, despite its tendency to suppress informative detail under the more stable depth conditions of S1.

The underlying mechanism remains uncertain. 
The present study does not reveal what specific aspects of the S2 eye-tracking signal are being altered by filtering or why these alterations lead to improved accuracy under variable calibration depth. 
Additional work—particularly analyses that examine how smoothing affects the spatial and temporal structure of gaze trajectories across different calibration depths—will be required to explain this asymmetric behavior. 
What the results make clear is that filtering is not universally beneficial or harmful; its impact depends strongly on the calibration conditions under which the data are collected, and this dependency warrants deeper investigation.

\subsection{Why does the \emph{Old} pipeline outperform the \emph{New} pipeline under the \emph{Both} axis condition?}

Across most configurations where an individual axis (optical or visual) was used, the \emph{New} pipeline—characterized by better spatial accuracy and precision—consistently yielded better authentication performance.
This aligns with the intuitive expectation that improved gaze signal quality should translate into lower EER and FRR.
However, when both axes were used jointly as input to the model, the pattern reversed: the \emph{Old} pipeline led to slightly lower EER and markedly reduced FRR at the 0.002\% FAR. 
Since the authentication model is identical across conditions—and in all cases receive the same pair of inputs (optical and visual axes)—these differences must arise from how each pipeline shapes the joint distribution of the two gaze signals. \\
A plausible interpretation is that the \emph{New} pipeline enhances each axis individually but also increases similarity or correlation between them. In the \emph{Both} condition, the model benefits from complementary information across optical and visual axes. If the \emph{New} pipeline reduces not only noise but also the subtle, person-specific differences between the axes, the incremental discriminative value of the second axis diminishes. By contrast, the \emph{Old} pipeline—despite being noisier—may retain a more diverse or less correlated relationship between the two axes, providing richer cross-axis structure that the model can exploit during authentication. \\
These findings do not imply that hardware or environmental conditions should drive the selection of the authentication pipeline. 
Rather, they highlight that improving signal quality in isolation does not automatically improve discriminability when multiple gaze axes are used jointly. 
Understanding why the \emph{Both}-axis configuration exhibits degraded performance under higher-quality signals remains an empirical question for future work.

\subsection{Additional Scenario: Reusing Enrollment Calibration for Verification}

In addition to the two implemented authentication scenarios, we propose a third scenario that we call S3. In this setting, the enrolled user’s original calibration (performed during enrollment) is reused for the verification stage.
For genuine matches, this is the same calibration as in enrollment.
For impostor matches, this represents a realistic attack scenario: an impostor attempts to access the device without recalibrating eye-tracking on it, using the enrolled (authentic) user’s existing calibration parameters.
To explore the feasibility of this scenario, we conducted a small pilot study using a subset of 250 users from the GazePro dataset. 
The findings were largely unsurprising and aligned with expectations: S1 consistently performs best, whereas S2 and S3 are similarly worst performing.

For example, in case of experiment 8, S1 achieved an EER of 0.17\% and FRR$_{0.002\%}$ of 24.26\%, while S2 and S3 showed much worse performance (S2: 9.64\% / 96.39\%, S3: 12.16\% / 98.37\%).
In experiment 18, S1 again outperformed with an EER of 0.18\% and FRR$_{0.002\%}$ of 32.25\%, compared to S2: 35.15\% / 99.38\% and S3: 23.70\% / 99.59\%.

A direct apple-to-apple comparison with the two implemented scenarios is not possible at this stage due to differences in the number of subjects. 
However, this pilot provides groundwork for future research, where a more controlled study could establish a meaningful comparative analysis across all three scenarios.

\subsection{Limitation \& Future work}

This work relies on a proprietary in-house GazePro dataset, which restricts both reproducibility and external benchmarking.
Because the dataset cannot be publicly released, independent verification of the results is limited, and comparisons with other gaze-based authentication systems remain constrained. 
This represents a substantive limitation of the present study.
The study also does not examine long-term feature permanence. 
Moreover, the short test–retest interval used here precludes any form of longitudinal evaluation.
Nevertheless, the baseline study on gaze-driven authentication performance conducted using the same dataset \cite{lohr2024baseline} provides a useful internal point of reference for comparison and address feature permanence.

Another limitation of the paper is external validation. We have thoroughly investigated the availability of publicly accessible eye-tracking datasets that would be suitable for external validation of our proposed method. Unfortunately, after an extensive search, we have found that there are no publicly available datasets that contain comparable eye-tracking data collected under similar conditions to our proprietary GazePro dataset. 

It is important to reiterate that, traditionally, the term ``optical axis'' denotes a purely anatomical reference line within the eye’s optical system. 
However, in the context of our paper, we use ``optical axis'' to refer to gaze directions estimated directly from our appearance-based gaze estimation pipeline—specifically, these are the raw gaze predictions generated before any calibration procedures.

Our RQ3 findings indicate that future research should explore why enhancements in spatial accuracy and precision can, at times, lead to decreased GA performance. 
This pattern suggests that the reliability of angle kappa estimation—previously shown to improve authentication accuracy~\cite{lohr2024baseline}—may be diminished. 
While this issue should be investigated further for the \emph{New} pipeline, it is beyond the scope of this paper.
It is also important to explore whether optimal multi-axis fusion depends on tailored noise calibration rather than simply enhancing signal quality.

Likewise, the findings for RQ6 pose an empirical question regarding why filtering yields improved performance under S2 conditions. 
This pattern suggests interactions between filtering and calibration depth that are not yet well understood and therefore warrant further investigation.

\section{CONCLUSION}

In this study, we systematically investigated the influence of critical factors on the performance of our state-of-the-art eye movement-based authentication method. 
Specifically, we examined the necessity of calibration and recalibration, the effects of calibration parameters during the training phase, signal quality, and training depth, as well as axis prioritization and the role of filtering.
Our experiments were conducted on a large-scale, proprietary dataset comprising nearly nine thousand subjects to rigorously evaluate authentication performance and elucidate the effects of each factor. 
We employed the state-of-the-art EKYT architecture, which ensures strong generalizability across diverse individuals.
Our findings showed that re/calibration is essential when calibration parameters change due to environmental factors, device repositioning, or even changes in target depth without shifting the device, as such depth changes alter the units of information encoded in the spatial-temporal domain of the ET signal when using an end-to-end gaze estimation pipeline.
We observed that model generalization remains robust, with no substantial change in authentication performance across different calibration parameters. 
While better spatial accuracy and precision generally yielded improved authentication, some mixed results were observed, suggesting the need for further investigation.
The best-ever performance achieved (EER of 0.01\% and an FRR$_{0.002\%}$ of 0.87\%) was obtained by combining optical and visual axes in a longer training depth setting.
Finally, the effect of filtering varies by scenario, sometimes improving and sometimes degrading authentication results.
These comprehensive investigations yielded robust and generalizable results, demonstrating the feasibility of deploying eye movement-based authentication systems in real-world settings.

\bibliographystyle{unsrt}
\bibliography{ms.bib}

\end{document}


\title{Supplementary Material for \textit{Gaze Authentication: Factors Influencing Authentication Performance}}

\author{
    Dillon~Lohr\textsuperscript{1}, 
    Michael J. Proulx\textsuperscript{1}, 
    Mehedi Hasan Raju\textsuperscript{2},  
    Oleg V. Komogortsev\textsuperscript{1,2}\\
    \textsuperscript{1}Meta Reality Labs Research, Redmond, WA, USA, 
    \textsuperscript{2}Texas State University, San Marcos, Texas, USA \\
    {\tt\small dlohr@meta.com, michaelproulx@meta.com, m.raju@txstate.edu, ok@txstate.edu}
}

\maketitle
\thispagestyle{empty}

\section*{Purpose of this document}
This supplementary material provides additional details about the quality of the signal from two different pipelines, \textit{Old} and \textit{New}.

\section{Spatial Accuracy and Precision Visualizations}
The main manuscript reports numerical summaries of binocular spatial accuracy (excluding fixations with dispersion $\geq$ 10 degrees) and spatial precision (S2S RMS) for the New and Old pipelines at target depths of 75 cm and 200 cm. All metrics are reported as the median of medians.  
This supplementary document provides the corresponding visualizations of these metrics, organized by pipeline and target depth as described below.

\begin{figure*}[htbp]
\centering
\includegraphics[width=0.9\textwidth]{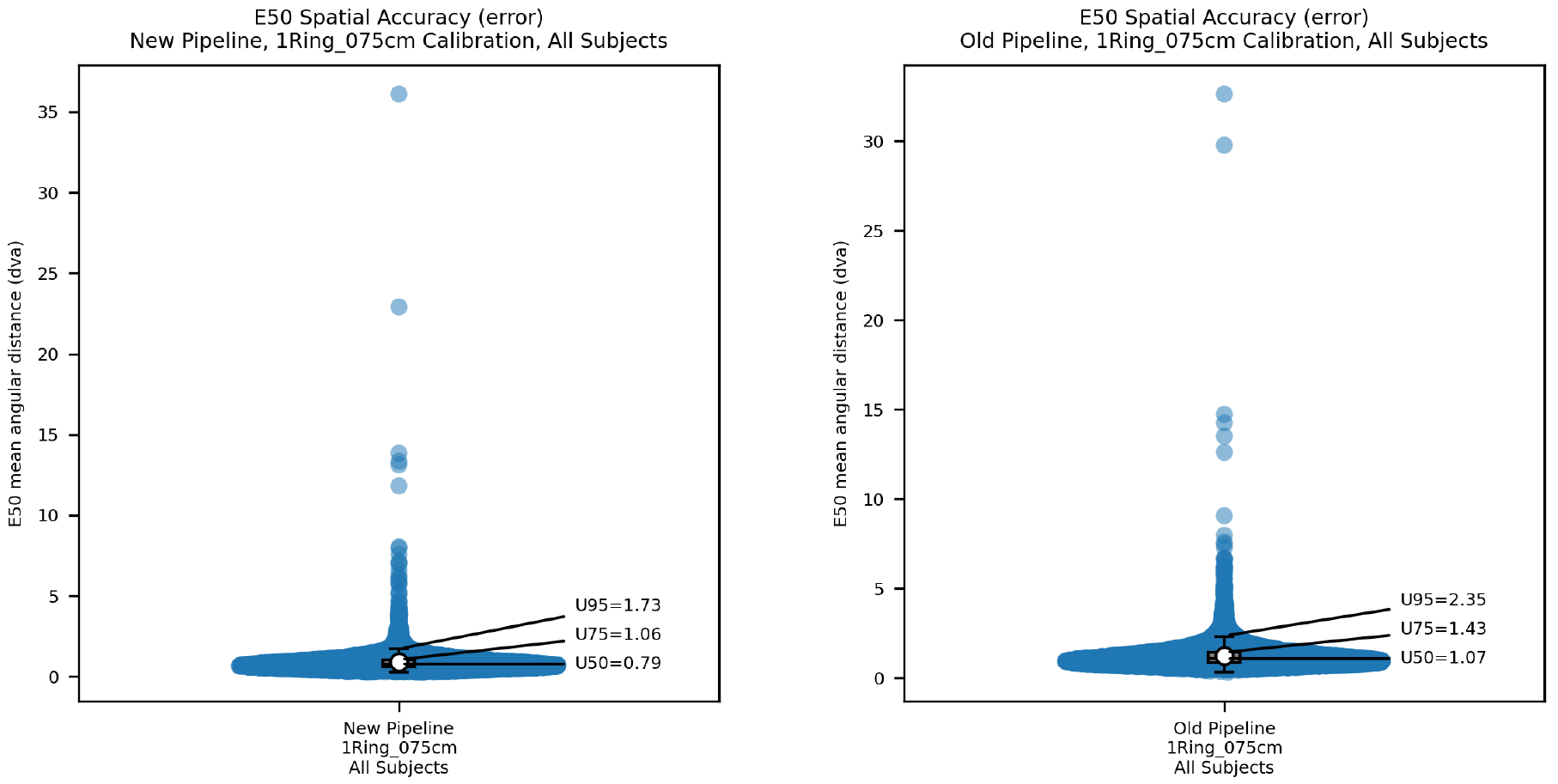}
\caption{E50 Binocular Spatial Accuracy for \textit{New} (left) and \textit{old} (right) pipeline with calibration at 75cm.}
\label{fig:s_75}
\end{figure*}

\begin{figure*}[htbp]
\centering
\includegraphics[width=0.9\textwidth]{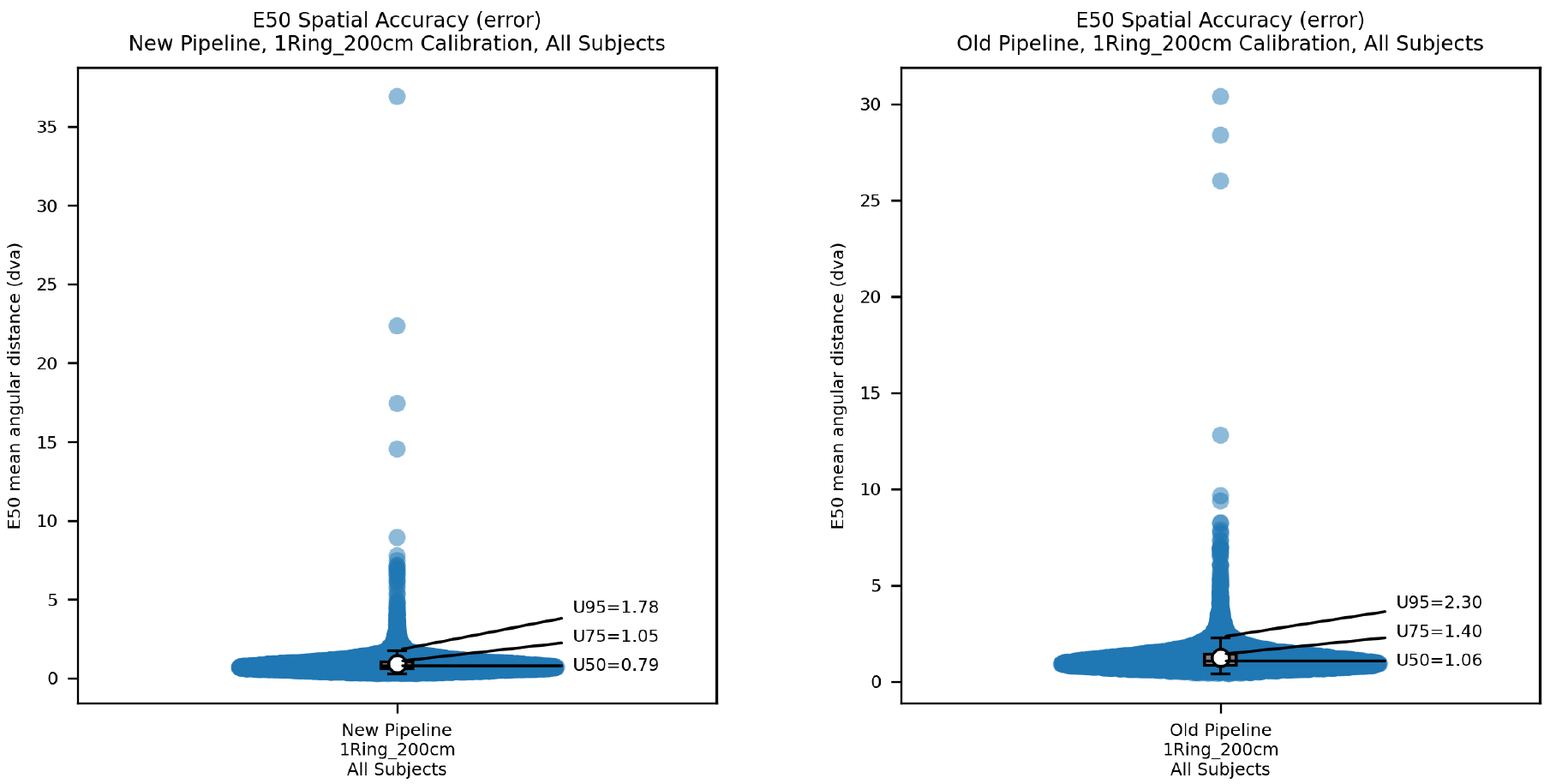}
\caption{E50 Binocular Spatial Accuracy for \textit{New} (left) and \textit{old} (right) pipeline with calibration at 200cm.}
\label{fig:s_200}
\end{figure*}

\begin{figure*}[htbp]
\centering
\includegraphics[width=0.9\textwidth]{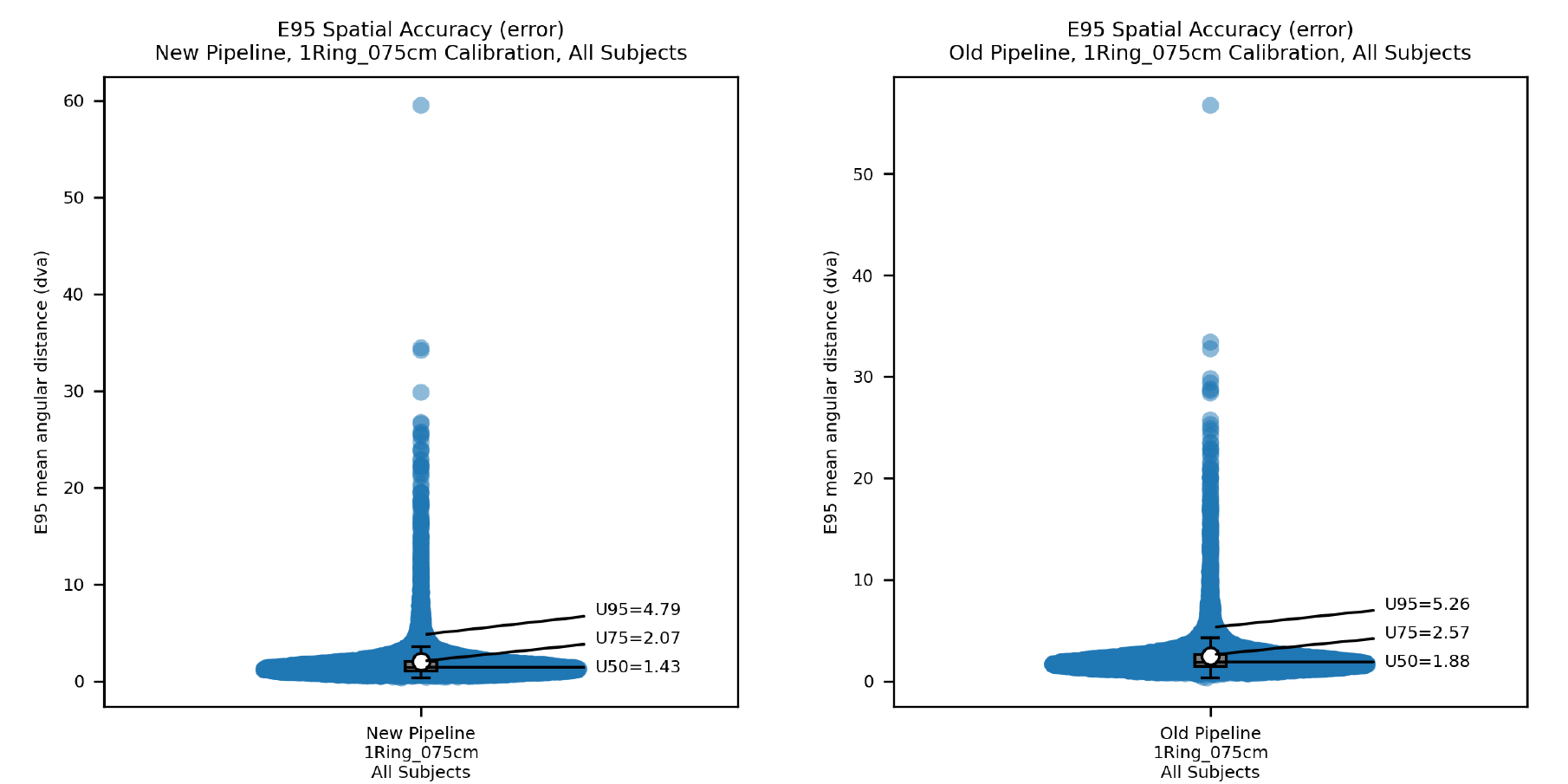}
\caption{E95 Binocular Spatial Accuracy for \textit{New} (left) and \textit{old} (right) pipeline with calibration at 75cm.}
\label{fig:s95_75}
\end{figure*}

\begin{figure*}[htbp]
\centering
\includegraphics[width=0.9\textwidth]{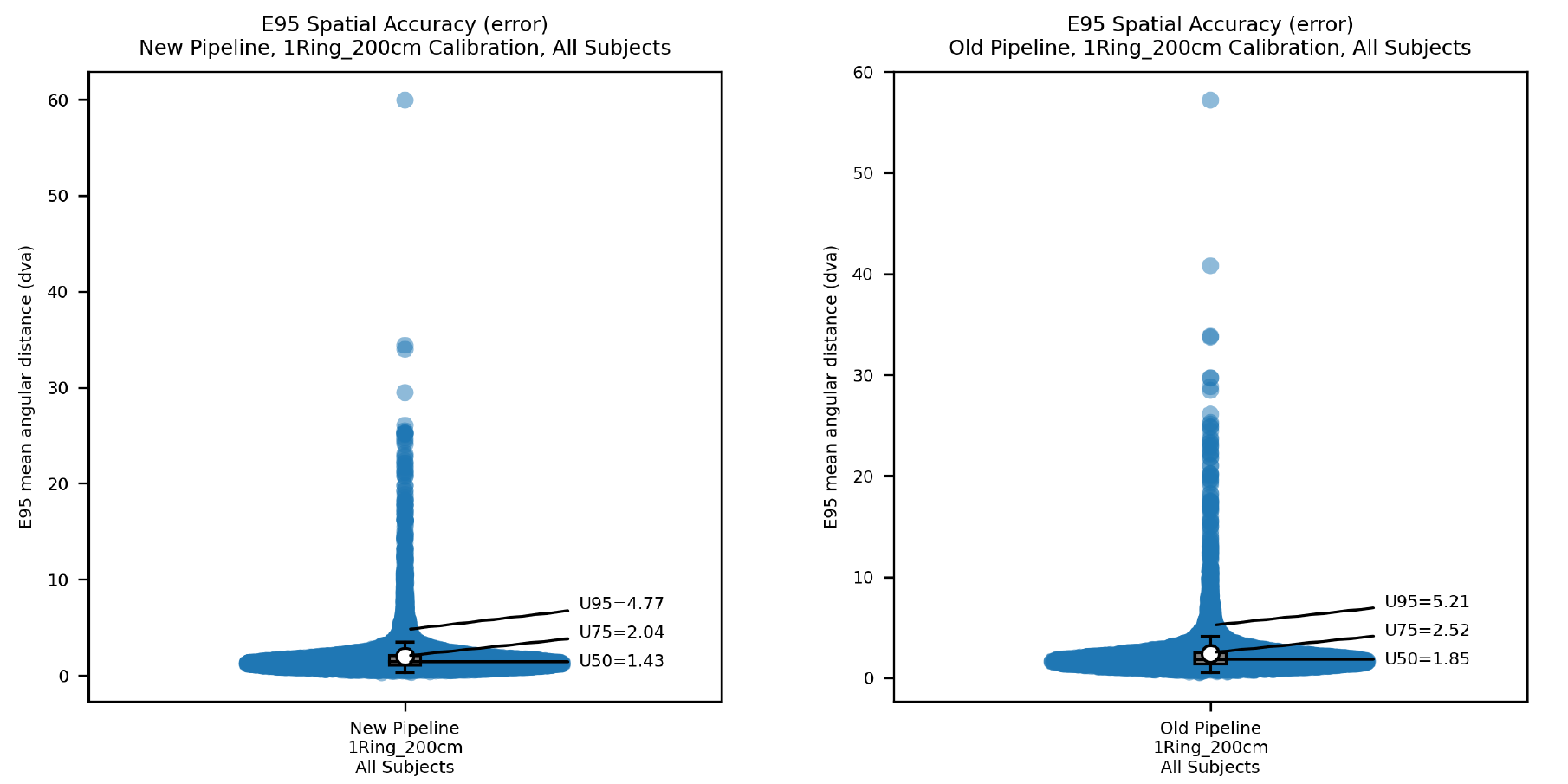}
\caption{E95 Binocular Spatial Accuracy for \textit{New} (left) and \textit{old} (right) pipeline with calibration at 200cm.}
\label{fig:s95_200}
\end{figure*}

\begin{figure*}[htbp]
\centering
\includegraphics[width=0.9\textwidth]{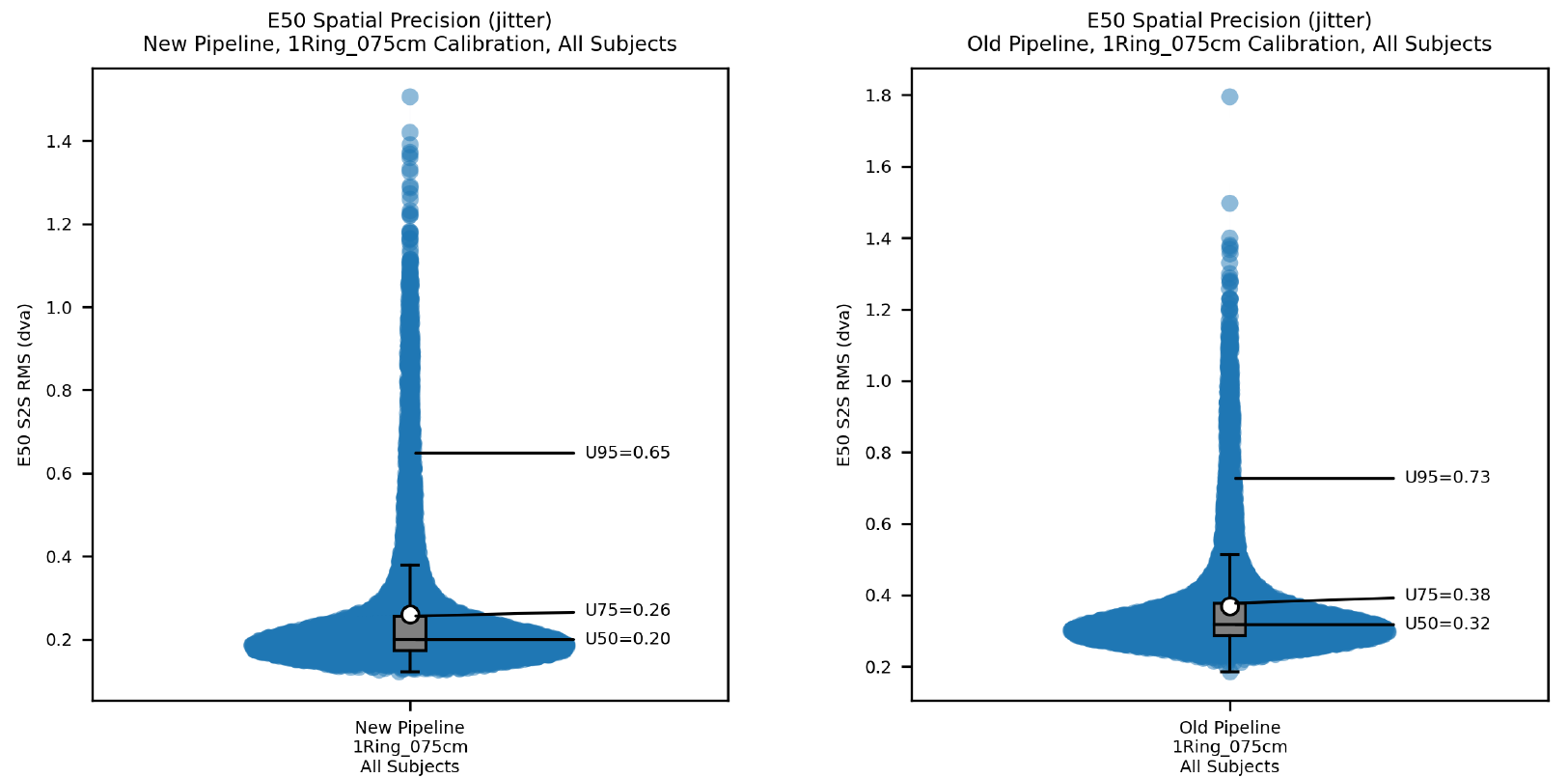}
\caption{E50 Binocular Spatial Precision (S2S RMS) for \textit{New} (left) and \textit{old} (right) pipeline with calibration at 75cm.}
\label{fig:p_75}
\end{figure*}

\begin{figure*}[htbp]
\centering
\includegraphics[width=0.9\textwidth]{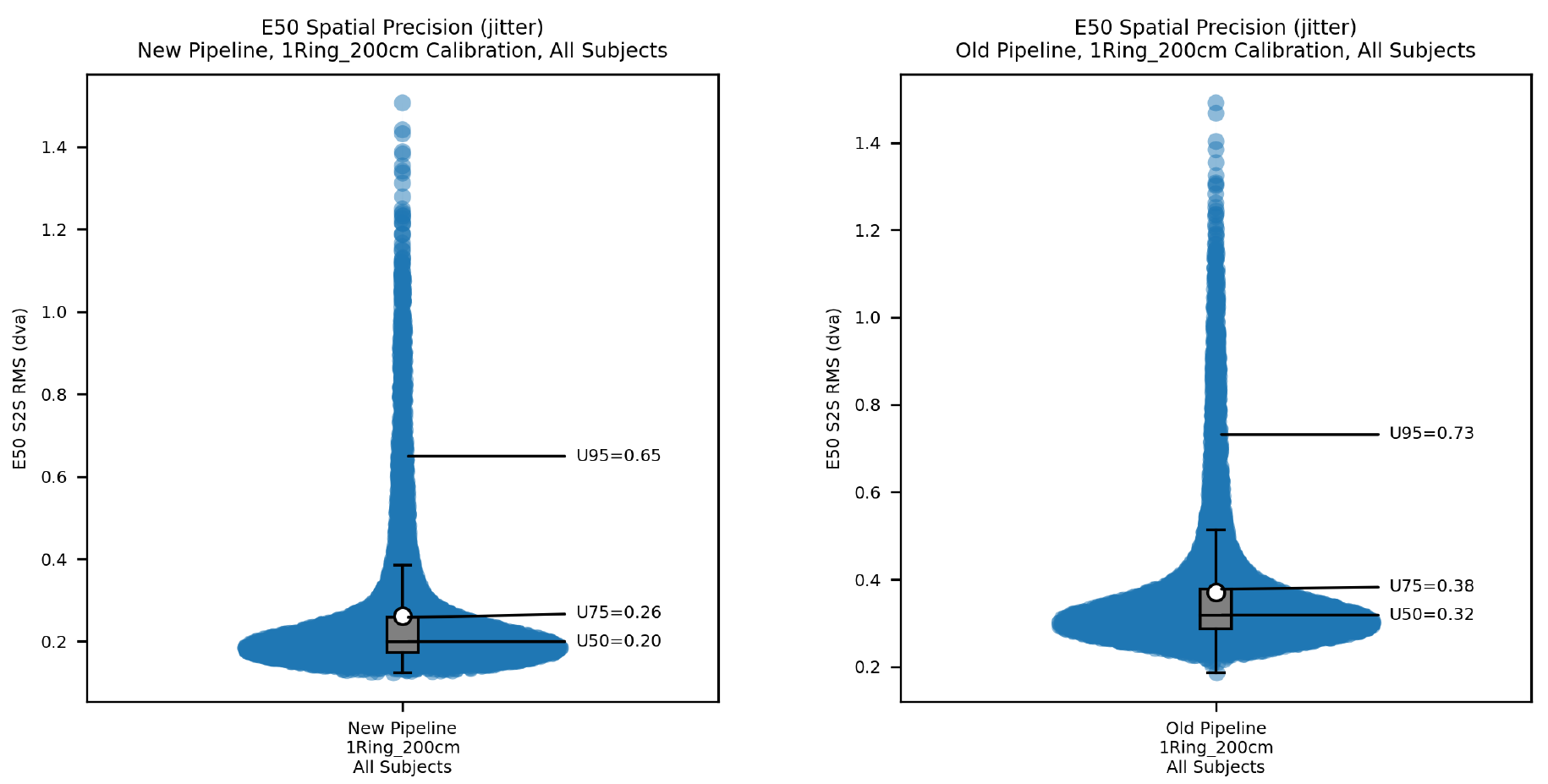}
\caption{E50 Binocular Spatial Precision (S2S RMS) for \textit{New} (left) and \textit{old} (right) pipeline with calibration at 200cm.}
\label{fig:p_200}
\end{figure*}

\begin{figure*}[htbp]
\centering
\includegraphics[width=0.9\textwidth]{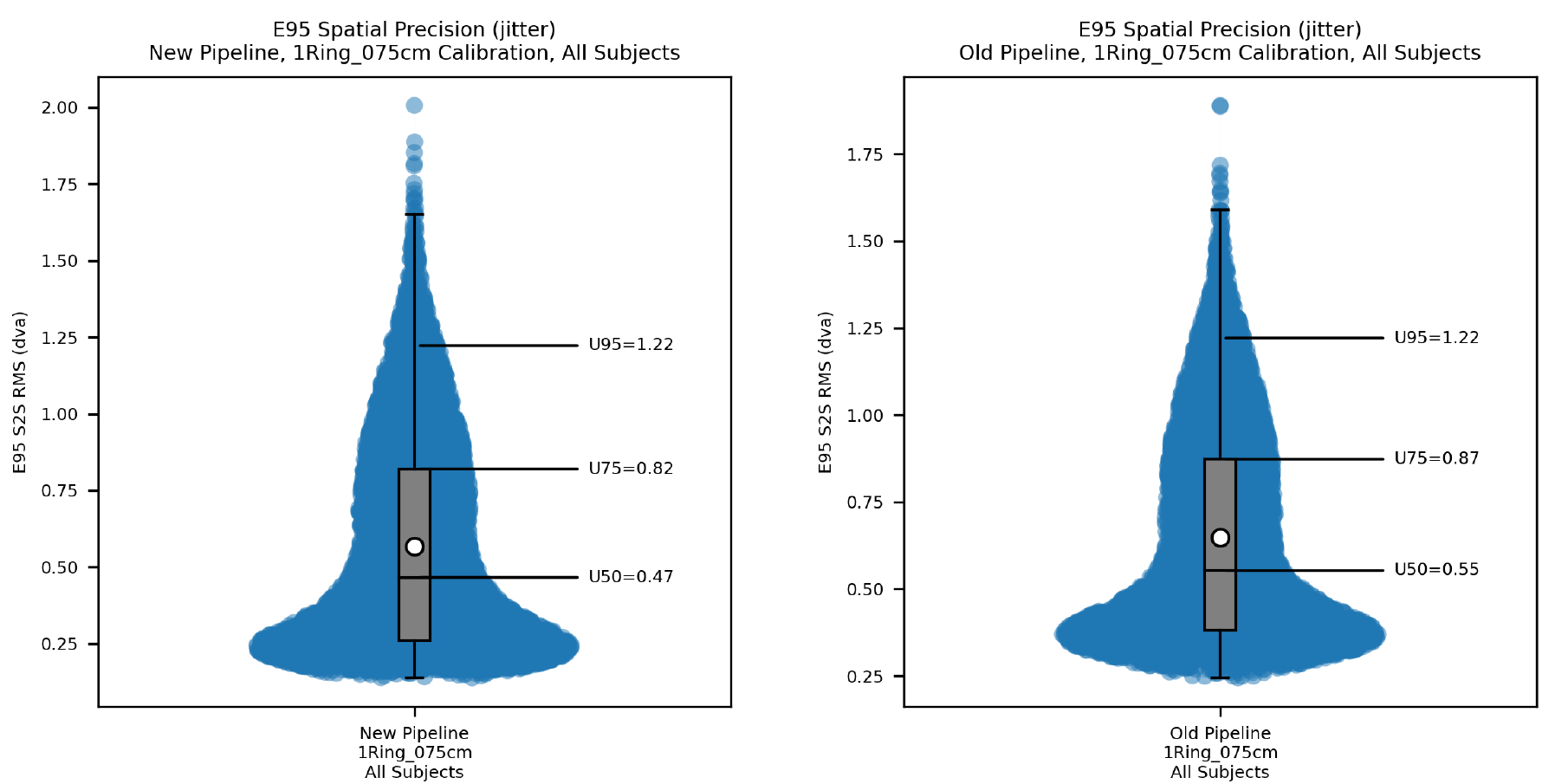}
\caption{E95 Binocular Spatial Precision (S2S RMS) for \textit{New} (left) and \textit{old} (right) pipeline with calibration at 75cm.}
\label{fig:p_75}
\end{figure*}

\begin{figure*}[htbp]
\centering
\includegraphics[width=0.9\textwidth]{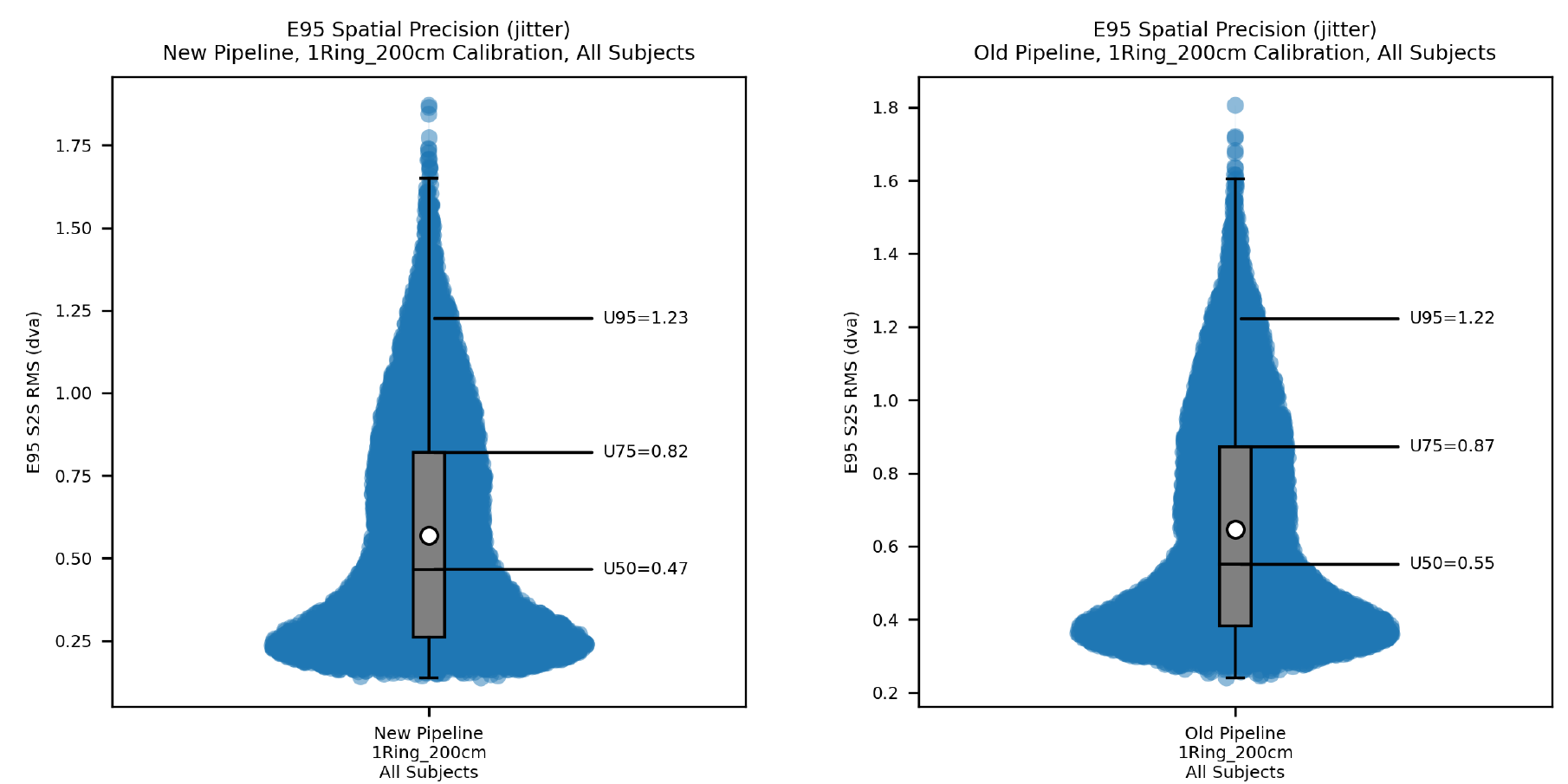}
\caption{E95 Binocular Spatial Precision (S2S RMS) for \textit{New} (left) and \textit{old} (right) pipeline with calibration at 200cm.}
\label{fig:p_200}
\end{figure*}

\bibliographystyle{unsrt}